\documentclass{article}

\usepackage[utf8]{inputenc}
\usepackage[T1]{fontenc}
\usepackage[breaklinks=true]{hyperref}
\usepackage{url}
\usepackage{booktabs}
\usepackage{amsfonts}
\usepackage{microtype}
\usepackage[table]{xcolor}
\usepackage{amsmath}
\usepackage{amssymb}
\usepackage{graphicx}
\usepackage{subcaption}
\usepackage{float}
\usepackage{placeins}
\usepackage[preprint]{icml2026}
\usepackage{multirow}

\makeatletter
\renewcommand{\Notice@String}{ICML 2026 Workshop.}
\makeatother

\graphicspath{{./}{../}}

\icmltitlerunning{Pretrained Model Representations as Acquisition Signals for Active Learning of MLIPs}

\begin{document}

\twocolumn[
\icmltitle{Pretrained Model Representations as Acquisition Signals for Active Learning of MLIPs}

\begin{icmlauthorlist}
\icmlauthor{Eszter Varga-Umbrich}{id,uni}
\icmlauthor{Shikha Surana$^\dagger$}{id}
\icmlauthor{Paul Duckworth}{id}
\icmlauthor{Jules Tilly}{id}
\icmlauthor{Olivier Peltre}{id}
\icmlauthor{Zachary Weller-Davies$^\dagger$}{id}
\end{icmlauthorlist}
\icmlaffiliation{id}{InstaDeep}
\icmlaffiliation{uni}{University of Cambridge}

\icmlcorrespondingauthor{Zachary Weller-Davies}{z.weller-davies@instadeep.com}

\vskip 0.3in
]

\renewcommand\thefootnote{}\footnotetext{\hspace{-0.45em}$^\dagger$Equal supervision.}\renewcommand\thefootnote{\arabic{footnote}}

\makeatletter
\def\@notice{}
\def\Notice@String{}
\makeatother

\printAffiliationsAndNotice{}

\begin{abstract}

Training machine learning interatomic potentials (MLIPs) for reactive chemistry is often bottlenecked by the high cost of quantum chemical labels and the scarcity of transition state configurations in candidate pools. Active learning (AL) can mitigate these costs, but its effectiveness hinges on the acquisition rule. We investigate whether the latent space of a pretrained MLIP already contains the information necessary for effective acquisition, eliminating the need for auxiliary uncertainty heads, Bayesian training and fine-tuning, or committee ensembles. We introduce two acquisition signals derived directly from a pretrained MACE potential: a finite-width neural tangent kernel (NTK) and an activation kernel built from hidden latent space features. On reactive-chemistry benchmarks, both kernels consistently outperform fixed-descriptor baselines, committee disagreement, and random acquisition, reducing the data required to reach performance targets by an average of 38\% for energy error and 28\% for force error. We further show that the pretrained model induces similarity spaces that preserve chemically meaningful structure and provide more reliable residual uncertainty estimates than randomly initialised or fixed-descriptor-based kernels. Our results suggest that pretraining aligns latent-space geometry with model error, yielding a practical and sufficient acquisition signal for reactive MLIP fine-tuning.

\end{abstract}

\section{Introduction}

Machine-learning interatomic potentials (MLIPs) have become practical surrogates for quantum chemical calculations, enabling simulations at far lower cost than density functional theory (DFT) \cite{Kohn1996DFT} while retaining useful accuracy across molecular and materials systems~\cite{Jacobs_2025,WANG2024109673,jmi.2025.17,behler2007generalized,batatia2023macehigherorderequivariant}.
Recent atomistic foundation models improve this further by learning reusable chemical and geometric representations from large datasets \cite{batatia2023macemp,wood2026umafamilyuniversalmodels,deng2023chgnet,
chen2022m3gnet}. However, even with strong pretraining, out-of-distribution (OOD) chemistry presents a fundamental bottleneck: accurate energies and forces require expensive reference calculations, and the model must be adapted to the specific chemical space of interest. Reactive chemistry, with its scarce transition-state configurations, is a canonical example of this challenge.

Active learning (AL) is a principled way to reduce this labelling burden
\cite{Settles2012}. In each round, a model selects a small batch of unlabelled
structures via an acquisition rule, obtains reference labels, and is fine-tuned on the enlarged training set. Sampling relevant unlabelled structures in reactive chemistry is itself difficult: molecular dynamics (MD) oversamples near-equilibrium configurations and undersamples transition states. More advanced samplers, such as metadynamics~\cite{Laio2002}, umbrella sampling~\cite{Torrie1977}, nudged elastic band (NEB) methods~\cite{Jonsson1998,Henkelman2000,Henkelman2000-2}, uncertainty-biased MD~\cite{zaverkin2024uncertaintybiased}, improve rare-event coverage but still encode the choices and limitations of the generator, and often require a potential accurate enough to support the exploration. 
Whichever generator is used, the resulting candidate pool is biased, and one must still decide which structures to label to achieve good performance on the target test distribution. We therefore study AL in the offline, pool-based setting, where a fixed candidate pool is given, and the acquisition rule alone determines labelling efficiency under this bias.

In the literature, AL for MLIPs often relies on committee disagreement or extrapolation criteria during simulation~\cite{smith2018less,podryabinkin2017active,jinnouchi2019mlff, vandermause2020flare,zhang2019uniformly, Achar2025, khan2026activelearningstrategiesefficient}. In contrast, representation-based batch AL treats selection as a geometry problem: construct a similarity space, then select structures that are uncertain or poorly covered in that space \cite{zaverkin2022batch_conformations,holzmuller2023batch_al}.

The central hypothesis of this work is that the latent space of a pretrained MLIP model already encodes sufficient information about model uncertainty for effective AL, removing the need for explicit uncertainty heads \cite{neumann2025orbv3,ho2025flexibleuncertaintycalibrationmachinelearned}, Bayesian training and fine-tuning \cite{jinnouchi2019mlff,vandermause2020flare,coscia2026blipsbayesianlearnedinteratomic}, or committee ensembles \cite{smith2018less,schran2020committee,peterson2017uncertainty,Kahle_2022}. 

We test this hypothesis primarily in pool-based active offline learning for reactive chemistry using pretrained MACE models \cite{batatia2023macehigherorderequivariant}. We introduce two model-based similarity metrics for AL: a finite-width energy neural tangent kernel (NTK) and a similarity kernel derived from hidden activation features. To our knowledge, finite-width NTK acquisition has previously been studied only for invariant-descriptor atomistic networks (Gaussian moment NNs)~\cite{zaverkin2022batch_conformations}, and was not adapted to SO(3)-equivariant pretrained MLIPs prior to the present work. Similarly, latent-space active learning has appeared
only in a narrow materials setting \cite{Ouyang2024} rather than in the context of
modern pretrained foundation models. We compare these methods against fixed-descriptor baselines such as SOAP (smooth overlap of atomic positions) \cite{bartok2013representing,de2017mlunifies,dscribe}, Morgan fingerprints \cite{Morgan1965,rogers2010extended,Ralaivola2005}, as well as committee disagreement and random acquisition. 

We evaluate primarily on the Transition1x (T1x) dataset~\cite{Schreiner2022}, a reactive-chemistry benchmark containing DFT energies and forces along NEB reaction pathways, with additional evaluation on the RGD~\cite{Zhao2023} and PMechDB~\cite{Tavakoli2024} reactivity subsets of OMol~\cite{levine2026openmolecules2025omol25}. Our main empirical finding is that pretrained model-based kernels are the most effective acquisition signals among the methods evaluated in the reactive settings we study, and they consistently outperform random acquisition, fixed-descriptor baselines, and committee ensembles (see Figure \ref{fig:natural-t1x} and Table~\ref{tab:transfer_benchmarks}). In particular, the NTK kernel coupled with the largest cluster maximum distance (LCMD) batch acquisition strategy reduces the number of acquisition rounds needed to reach a shared target by an average of \(38.1\%\), \(28.3\%\), \(27.2\%\), and \(8.3\%\) for energy RMSE, force RMSE, energy MAE, and force MAE, respectively. 

In our T1x case study, the accompanying diagnostics support a mechanistic interpretation: compared with randomly initialised neural kernels and fixed descriptors, pretrained model-based kernels yield better residual interpolation, better uncertainty calibration, and a similarity geometry that preserves both reaction-family structure and variation along reaction paths. Although we focus on a specific pretrained MACE model, our results suggest that pretrained model-based representations already encode uncertainty-relevant structure, can be used effectively for active learning, and merit broader study as practical acquisition signals for MLIPs.
 
\section{Related Work}
\paragraph{Active learning for MLIPs.}
Active learning has become a standard tool for reducing the cost of training MLIPs. Most prior work has focused on \emph{online} active learning, in which candidate generation and acquisition are coupled: structures are generated on the fly and selected using uncertainty or extrapolation criteria. Representative examples include committee-based approaches~\cite{smith2018less,schran2020committee}, extrapolation-based methods for moment-tensor potentials~\cite{podryabinkin2017active}, Bayesian force-field models~\cite{jinnouchi2019mlff,vandermause2020flare, coscia2026blipsbayesianlearnedinteratomic}, concurrent learning~\cite{zhang2019uniformly}, and uncertainty-driven dynamics~\cite{kulichenko2023uncertainty}. In contrast, offline pool-based active learning operates on a fixed set of candidate structures~\cite{zaverkin2022batch_conformations,DDP2026}, where acquisition must contend with both scale and distribution shift. This is particularly pronounced in reactive chemistry, where candidate pools are often strongly biased relative to the target distribution~\cite{deng2025softening,cui2025tta}.

\paragraph{Pretrained representations as acquisition signals.}
Finite-width energy NTKs have previously been studied as active-learning signals in Gaussian Moment Networks \cite{zaverkin2022batch_conformations}, which use invariant descriptors as inputs to a feedforward MLP. Other methods \cite{holzmuller2023batch_al} provide a broader benchmark for deep batch active learning in the regression setting. Likewise, uncertainty quantification based on hidden latent-space representations has been explored in Gaussian Moment Networks \cite{zhu2023fast_uq}, and related latent-space active learning has appeared in prior work \cite{Ouyang2024}, but only in a relatively narrow HfO$_2$ materials setting rather than in the context of modern pretrained foundation models. By contrast, fixed descriptor spaces such as SOAP~\cite{bartok2013representing,de2017mlunifies} and fingerprint similarities~\cite{rogers2010extended,Ralaivola2005} provide model-independent notions of structural similarity and have also been used for atomistic data selection~\cite{DDP2026}. More recently, uncertainty in MLIPs has also been approached through uncertainty-calibration or confidence heads \cite{Tan_2023,ho2025flexibleuncertaintycalibrationmachinelearned,neumann2025orbv3}, Bayesian networks \cite{coscia2026blipsbayesianlearnedinteratomic}, and ensemble-based uncertainty estimates. While committee-based uncertainty has been studied extensively in MLIPs, obtaining committee-style uncertainty from a pretrained model remains relatively underexplored. A notable recent exception is the multi-head committee approach of \cite{beck2025multihead}. Our work asks a different question: whether, for \textit{pretrained} equivariant foundation models, uncertainty-relevant acquisition geometry can be extracted directly from the pretrained representation itself.

\section{Methods}

\subsection{Offline Active Learning}

We study pool-based active learning for machine-learned interatomic potentials. At round $t$, the learner has a labelled training set $\mathcal{T}^{(t)}$ and a fixed unlabelled candidate pool $\mathcal{P}^{(t)}$ containing $n_{\mathcal{P}}$ points. Here, a \textit{candidate pool} is an unlabelled set of potential objects on which one can evaluate and decide whether to acquire the label. Each structure in the candidate pool is denoted by
$\mathbf{x}=(\mathbf{z},\mathbf{r})$, with atomic
numbers $\mathbf{z}=(z_i)_{i=1}^{N(\mathbf{x})}$ and Cartesian coordinates
$\mathbf{r}=(\mathbf{r}_i)_{i=1}^{N(\mathbf{x})}$, 
where $N(\mathbf{x})$ is the number of atoms in the structure.

The goal of a \textit{batch acquisition rule} is to select the most informative candidates to label to efficiently improve downstream test error. For the case of MLIPs, a label for a structure $\mathbf{x}$ is the associated energies and forces of a DFT calculation
\begin{equation}
y
=
\left(
E(\mathbf{x}),
\{\mathbf{F}_i(\mathbf{x})\}_{i=1}^{N(\mathbf{x})}
\right),
\end{equation}
where $E(\mathbf{x}) \in \mathbb{R}$ is the total energy and $\mathbf{F}_i(\mathbf{x}) \in \mathbb{R}^3$ is the force on atom $i$, given by $\mathbf{F}_i(\mathbf{x})\label{eq: forces}=-\nabla_{\mathbf{r}_i} E(\mathbf{x})$.

Formally, a batch acquisition rule selects $\mathcal{A}^{(t)} \subseteq \mathcal{P}^{(t)}$ with $|\mathcal{A}^{(t)}| = B$. Reference labels are then revealed and the training and candidate pool sets are updated as
\begin{equation}
\mathcal{T}^{(t+1)}
=
\mathcal{T}^{(t)}
\cup
\{(\mathbf{x},y) : \mathbf{x}\in\mathcal{A}^{(t)}\},
\end{equation}
and
$\mathcal{P}^{(t+1)}
=
\mathcal{P}^{(t)} \setminus \mathcal{A}^{(t)}$.

The model is fine-tuned after each acquisition round on all of $\mathcal{T}^{(t+1)}$. All experiments use the MACE architecture~\cite{batatia2023macehigherorderequivariant} through a fork of the \texttt{mlip} library~\cite{brunken2025machinelearninginteratomicpotentials}. The MACE architecture is a parameterised map taking atomic structures to energy predictions $E_{\theta}(\mathbf{x})$, where $\theta$ denotes the parameters of the network; forces are obtained by auto-differentiation.

Models are initialised from the same internally trained SPICE-2~\cite{SPICE, levine2026openmolecules2025omol25} model; all labels are at the $\omega$B97M-V/def2-TZVPD level of theory used in OMol25~\cite{levine2026openmolecules2025omol25}. Further architecture and training details are given in Appendices~\ref{app:mace-architecture} and~\ref{app:training-details}.

\subsection{Acquisition Signals}

The purpose of an acquisition signal is to rank candidates in the unlabelled pool and select those that are expected to most improve the model. We distinguish between two broad classes of acquisition methods. 

\paragraph{Kernel-based methods:} These imply constructing a similarity kernel over structures $k(\mathbf{x}, \mathbf{x}')$, typically from a structure-level embedding $\phi(\mathbf{x}) \in \mathbb{R}^d$, where $d$ is the embedding dimension, and then apply a downstream batch selection rule such as posterior variance. Within the kernel-based family, we further distinguish between \emph{model-based representations}, extracted from the pretrained force field itself, and \emph{feature-based representations}, built from fixed molecular descriptors. 

\paragraph{Direct uncertainty methods: } These approaches assign acquisition scores directly, for example, through committee disagreement. Though a kernel can implicitly be used to quantify model-uncertainty, we think this distinction is important because a kernel provides pairwise information about similarity, coverage, and redundancy across the pool, whereas a direct uncertainty score is pointwise and does not by itself encode relations among candidates. 

The central question of this work is whether pretrained \emph{model-based representations} already contain information that is useful for AL. A summary of this taxonomy is given in Table~\ref{tab:acquisition_methods}, while Table~\ref{tab:cost_comparison} in Appendix~\ref{app:full-natural-bias} summarises the practical computational considerations for each method introduced in this section.

\begin{table}[t]
\scriptsize
\centering
\caption{Summary of acquisition methods considered in this work. Kernel-based methods first define a similarity kernel and then apply a batch selector such as posterior variance or LCMD. Direct methods assign acquisition scores without an intermediate kernel.}
\label{tab:acquisition_methods}
\begin{tabular}{lll}
\toprule
Method & Representation / signal & Score construction \\
\midrule
\multicolumn{3}{l}{\textbf{Model-based}} \\
Activation & Hidden activation features & Kernel-based \\
Energy NTK & Energy gradient features & Kernel-based \\
\midrule
\multicolumn{3}{l}{\textbf{Feature-based}} \\
SOAP & SOAP descriptor & Kernel-based \\
Tanimoto & Morgan fingerprint & Kernel-based (Tanimoto) \\
\midrule
\multicolumn{3}{l}{\textbf{Uncertainty-based}} \\
Committee-E & Energy disagreement & Direct score \\
Committee-F & Force disagreement & Direct score \\
\midrule
\multicolumn{3}{l}{\textbf{Baseline}} \\
Random & None & Uniform sampling \\
\bottomrule

\end{tabular}
\end{table}

\subsubsection{Kernel-Based Methods}

For kernel-based methods, structures are first mapped to a representation $\phi(\mathbf{x})$, from which we construct a similarity kernel. For continuous embeddings, we use the cosine-normalised kernel
\begin{equation}\label{eq: kernel_main_short}
k(\mathbf{x},\mathbf{x}')
=
\tilde{\phi}(\mathbf{x})^\top \tilde{\phi}(\mathbf{x}'),
\qquad
\tilde{\phi}(\mathbf{x})
=
\frac{\phi(\mathbf{x})}{\max(\|\phi(\mathbf{x})\|_2,\varepsilon)}.
\end{equation}
Equation~\eqref{eq: kernel_main_short} is what we use in practice to construct similarity kernels between structures. The exception is for Morgan fingerprints, which are binary set-like descriptors and are therefore compared using Tanimoto similarity,
\begin{equation}
k_{\mathrm{Tan}}(\mathbf{x},\mathbf{x}')
=
\frac{|\phi(\mathbf{x})\cap \phi(\mathbf{x}')|}
     {|\phi(\mathbf{x})\cup \phi(\mathbf{x}')|}.
\end{equation}

We now introduce two model-based representations used to extract representations from pretrained models.

\paragraph{\textit{Model-based representations}}

\paragraph{Energy NTK features:}
The finite-width energy neural tangent kernel represents a structure by the local sensitivity of the model energy prediction (in this instance, MACE) to parameter perturbations,
\begin{equation}
\phi_{\mathrm{NTK}}(\mathbf{x})
=
\nabla_{\theta} E_\theta(\mathbf{x}),
\end{equation}
where dimension of $\phi_{\mathrm{NTK}}$ is the number of parameters. Intuitively, two structures with similar NTK features cause similar perturbations to the model's predictions and therefore probe overlapping directions of parameter sensitivity. NTK feature similarity then becomes a natural measure of coverage between the candidate pool and the current training set, and a natural acquisition signal for how much a new structure would improve the model.

A key practical issue in pretrained networks is that one cannot reasonably form NTK features with respect to the entire parameter space: the resulting gradient representation is prohibitively large and too expensive for repeated scoring over a candidate pool~\cite{zaverkin2022batch_conformations}. It is therefore necessary to restrict the NTK to a parameter subspace $\theta_P$. In this work, we use the embedding parameter blocks of MACE for $\theta_P$ (see Appendix \ref{app:mace-architecture}). This subset captures the model's chemical encoders explicitly, and keeps feature extraction tractable for scoring. For the MACE model used in this work, this yields a 1920-dimensional feature vector. A broader study of reduced parameter subsets is left for future work.

\paragraph{Activation features:}
As a cheaper model-dependent representation, we extract hidden MACE activations. Let $h_i^{(\ell)}(\mathbf{x})$ be the atom-wise hidden state at interaction layer $\ell$. Because MACE features contain multiple rotation orders, we retain the invariant scalar channels and take the mean over atoms:
\begin{equation}
\phi_{\mathrm{act}}(\mathbf{x})
=
\operatorname*{concat}_{\ell}
\frac{1}{N(\mathbf{x})}\sum_{i=1}^{N(\mathbf{x})}
h_{i,L=0}^{(\ell)}(\mathbf{x}).
\end{equation}
All experiments use a two-layer MACE model with 128 scalar hidden channels per layer, giving a 256-dimensional activation vector. Activation features are much quicker to compute than the NTK: they require only a single forward pass, whereas NTK features require reverse-mode differentiation of the energy with respect to $\theta_P$. 

\paragraph{\textit{Feature-based representations}}

\paragraph{SOAP:}
Smooth Overlap of Atomic Positions (SOAP)~\cite{bartok2013representing} features are computed with \texttt{DScribe}~\cite{dscribe} using $r_{\mathrm{cut}}=6$~\AA, $n_{\max}=8$, and $l_{\max}=6$. Since SOAP features depend on the local atomic environment, we aggregate local SOAP descriptors to a structure-level representation by averaging over the power spectrum of different sites using the outer averaging mode of \texttt{DScribe}.

\paragraph{Morgan fingerprints:}
Morgan fingerprints~\cite{Morgan1965} use radius 3 and 2048 bits, and are paired with the Tanimoto kernel~\cite{Ralaivola2005}. We compute Morgan fingerprints using the open source library \texttt{RDKit} \cite{rdkit_2025_09_1}. They provide a cheap chemistry-only baseline, complementary to SOAP's local geometric descriptor and to the model-based MACE representations.

\subsubsection{Direct Uncertainty Methods}

\paragraph{Committee disagreement:}
Committee methods estimate uncertainty by training an ensemble of $M$ MACE models $\{f_{\theta_m}\}_{m=1}^M$ on the current labelled set and scoring candidates by prediction disagreement. For energy acquisition (Committee-E), we use the standard deviation of predicted energies,
\begin{equation}
\alpha_{\mathrm{com}}^{E}(\mathbf{x})
=
\operatorname{std}_{m=1}^M
E_{\theta_m}(\mathbf{x}).
\end{equation}
For force acquisition (Committee-F), we compute disagreement over force components. In our committee benchmarks, we use $M=3$ models initialised from the same pretrained checkpoint, with diversity introduced by independent data-order seeds of $\mathcal{T}^{(t)}$. Committee methods provide a direct uncertainty score but increase training and evaluation cost approximately linearly with $M$. Prior work employs small ensembles (typically 3–5 models \cite{Achar2025, kulichenko2023uncertainty, khan2026activelearningstrategiesefficient}), and we adopt $M=3$ as a minimal, cost-efficient configuration that yields a well-defined variance estimate for disagreement-based acquisition.

A practical complication in the pretrained-model setting is that meaningful ensemble diversity is difficult to obtain: all committee members begin from the same pretrained checkpoint, so diversity must be induced during fine-tuning rather than through independent pretraining. In our committee benchmarks in Appendix \ref{sec:appendix-committee-natural-bias}, we explore two such strategies: independent data-order shuffling and bootstrap resampling of the current labelled set $\mathcal{T}^{(t)}$. In expectation, each bootstrap sample contains about $63.2\%$ unique examples from the original dataset. On T1x, we find that, under this committee construction, the shuffle variant performs better, and it is therefore the version reported in the main text. For completeness, we also consider committees trained from scratch with different initialisation seeds (Appendix \ref{sec: scratch}) and observe qualitatively similar results. 

\subsection{Batch Selection Rules}

For kernel-based methods, the representation is first converted into a similarity kernel and then coupled to a batch selection rule. For direct uncertainty methods, such as committee disagreement, candidates are ranked directly by the acquisition score.

\paragraph{Posterior variance (PV):}
For kernel methods, the main selection rule is greedy Gaussian-posterior variance. Given a kernel $k$ and current training set $\mathcal{T}^{(t)}$, a candidate $\mathbf{x}$ has posterior variance
\begin{equation}
\sigma_t^2(\mathbf{x})
=
k(\mathbf{x},\mathbf{x})
-
k_{\mathcal{T}^{(t)}}(\mathbf{x})^\top
(k_{\mathcal{T}^{(t)}\mathcal{T}^{(t)}}+\lambda I)^{-1}
k_{\mathcal{T}^{(t)}}(\mathbf{x}),
\end{equation}
where $k_{\mathcal{T}^{(t)}}(\mathbf{x})$ is the vector of similarities between $\mathbf{x}$ and the training set. The batch is constructed greedily: after each selection, the newly selected point is added to the conditioning set before the next point is chosen~\cite{Rasmussen2005}.

\paragraph{LCMD:}
We also use kernel coverage rules. Largest-cluster maximum-distance (LCMD)~\cite{holzmuller2023batch_al} modifies greedy farthest-point sampling to favour dense but under-covered regions of the pool. Let $S$ denote the set of points already selected into the current batch. At each step, every remaining candidate is assigned to its nearest centre in $S \cup \mathcal{T}^{(t)}$. For each cluster $\mathcal{C}_c$ with center $\mathbf{z}_c$, we compute the total squared distance mass
\begin{equation}
m_c
=
\sum_{\mathbf{x} \in \mathcal{C}_c}
d_k^2(\mathbf{x},\mathbf{z}_c),
\end{equation}
where the kernel-induced squared distance is
\begin{equation}
d_k^2(\mathbf{x},\mathbf{z})
=
k(\mathbf{x},\mathbf{x})
+
k(\mathbf{z},\mathbf{z})
-
2k(\mathbf{x},\mathbf{z}).
\end{equation}
LCMD selects the cluster with the largest $m_c$ and then adds the candidate in that cluster farthest from its centre,
\begin{equation}
\mathbf{x}_{\mathrm{next}}
=
\arg\max_{\mathbf{x}\in\mathcal{C}_{c^\star}}
d_k^2(\mathbf{x},\mathbf{z}_{c^\star}),
\ \
c^\star = \arg\max_c m_c.
\end{equation}

\begin{figure*}[!t]
\centering
\begin{minipage}{\textwidth}
  \centering
  \includegraphics[width=\linewidth]{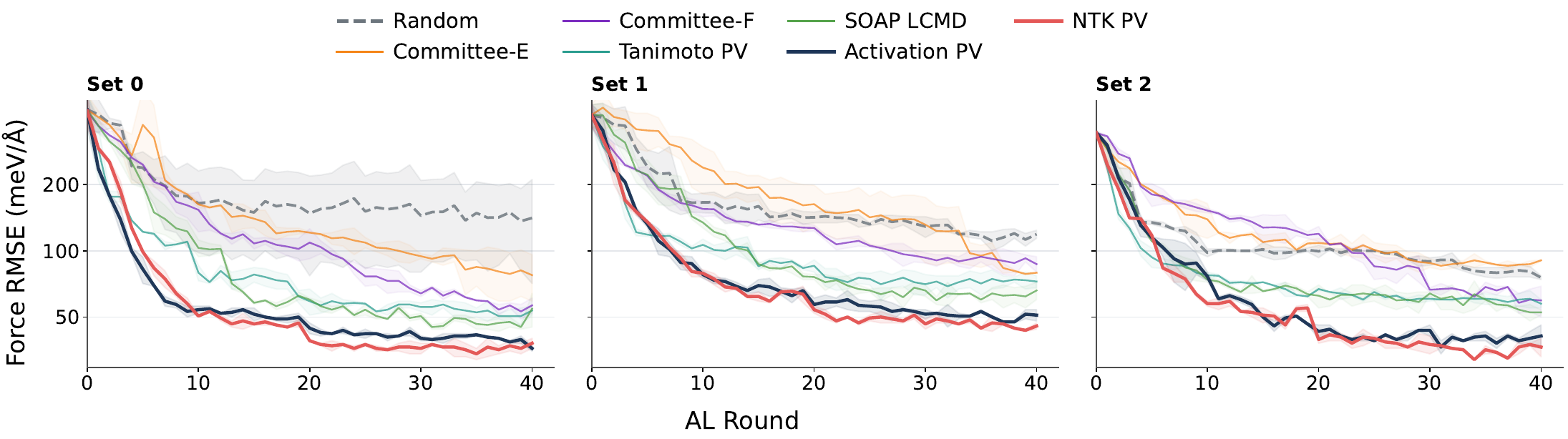}
\end{minipage}

\vspace{0.45em}

\begin{minipage}{\textwidth}
  \centering
  \scriptsize
    \setlength{\tabcolsep}{8pt}
\begin{tabular}{lcccc}
\toprule
Method & Energy AUC ($\downarrow$) & Force AUC ($\downarrow$) & Final Energy ($\downarrow$) & Final Force ($\downarrow$) \\
\midrule
Random          & 648.33          & 6490.00         & 6.33          & 111.83          \\
Committee-E     & 623.33          & 6636.67          & 3.50          & 82.50          \\
Committee-F     & 915.33          & 5496.00          & 8.00          & 67.67          \\
NTK (PV)        & 252.83          & \textbf{3078.00} & \textbf{2.83} & \textbf{40.00} \\
Activation (PV) & \textbf{252.17} & 3179.00          & 3.17          & 42.67          \\
Tanimoto (PV)   & 275.33          & 3791.83          & 3.17          & 61.17          \\
SOAP (LCMD)     & 278.83          & 4270.67          & 3.00          & 57.67          \\
\bottomrule
\end{tabular}

\end{minipage}
\caption{Force RMSE (meV\,\AA$^{-1}$) under the natural T1x setting. Each acquisition step adds
five structures. The table reports metrics averaged across the natural T1x pools;
lower is better for all columns. Among the methods compared here, model-dependent kernels are the strongest
acquisition signals: NTK-PV gives the best force AUC and final force error,
while Activation-PV gives the best energy AUC.}
\label{fig:natural-t1x}
\end{figure*}

\section{Results}

We build the case for pretrained model representations as acquisition signals in three steps: learning curves on a reactive-chemistry benchmark (Section \ref{sec: t1x_case_study}), followed by two complementary diagnostics, kernel geometry (Section \ref{sec:kernel-geometry-diagnostics}) and residual uncertainty calibration (Section \ref{sec:residual-gp-uncertainty-calib}), and finally generalisation to additional reactive datasets (Section \ref{sec:transferability}). Together, these results support a unified interpretation: \emph{pretraining shapes the latent space into a geometry that is already aligned with model error, making it a practical and sufficient acquisition signal}.

\subsection{T1x Case Study}\label{sec: t1x_case_study}
Transition1x (T1x) \cite{Schreiner2022} contains DFT energies and forces along NEB and climbing-image NEB reaction paths. Each reaction is represented as a sequence of frames connecting reactant, transition-state-like, and product configurations, so the frame index provides a discrete reaction coordinate. This makes T1x a useful benchmark for testing whether acquisition strategies can distinguish different reaction pathways while also resolving variation along the reaction coordinate.

We construct three candidate pools, labelled Set 0, Set 1, and Set 2, containing approximately 1.7k, 4.2k, and 5.7k structures, respectively. Each pool is built from five randomly selected T1x reaction pathways, and we retain the natural imbalance of the sets, so the number of structures per reaction may differ. At each acquisition round, the active-learning method adds five new structures to the training set. All methods are evaluated on the same fixed balanced test set, containing 35 structures from each reaction pathway. Each active-learning run is repeated with two different seeds.

Figure~\ref{fig:natural-t1x} summarises the central active-learning result on T1x. For each method, we report only its best-performing selection rule; a full breakdown of results is provided in Appendix~\ref{app:full-natural-bias}, and the overall pattern remains the same. We also compare against training from scratch in Appendix \ref{sec: scratch} and find the results are qualitatively the same, with all methods finding a consistent advantage in using a pretrained model.

Model-based kernel methods are the most sample-efficient acquisition signals.
Activation-PV gives the best energy area under the curve (AUC), and NTK-PV gives the best force AUC,
best final energy error, and best final force error among the reported methods.
Both learned kernels substantially outperform random selection. Descriptor methods are useful but weaker: Tanimoto-PV improves over random, and SOAP-LCMD is competitive on final energy, but neither matches the neural kernels on force learning. Committee disagreement is less reliable, with committee-energy worse than random by force AUC, and committee-force trading improved force error for much worse energy error.

The learning-curve advantage shows pretrained representations induce a useful similarity space. We now probe the kernel itself: how distinctively does it separate structurally different configurations, and how much fine-grained resolution does it preserve?

\subsection{Kernel Geometry Diagnostics}
\label{sec:kernel-geometry-diagnostics}

\begin{figure*}[t]
    \centering
    \begin{subfigure}[t]{0.32\textwidth}
      \centering
      \includegraphics[width=\linewidth]{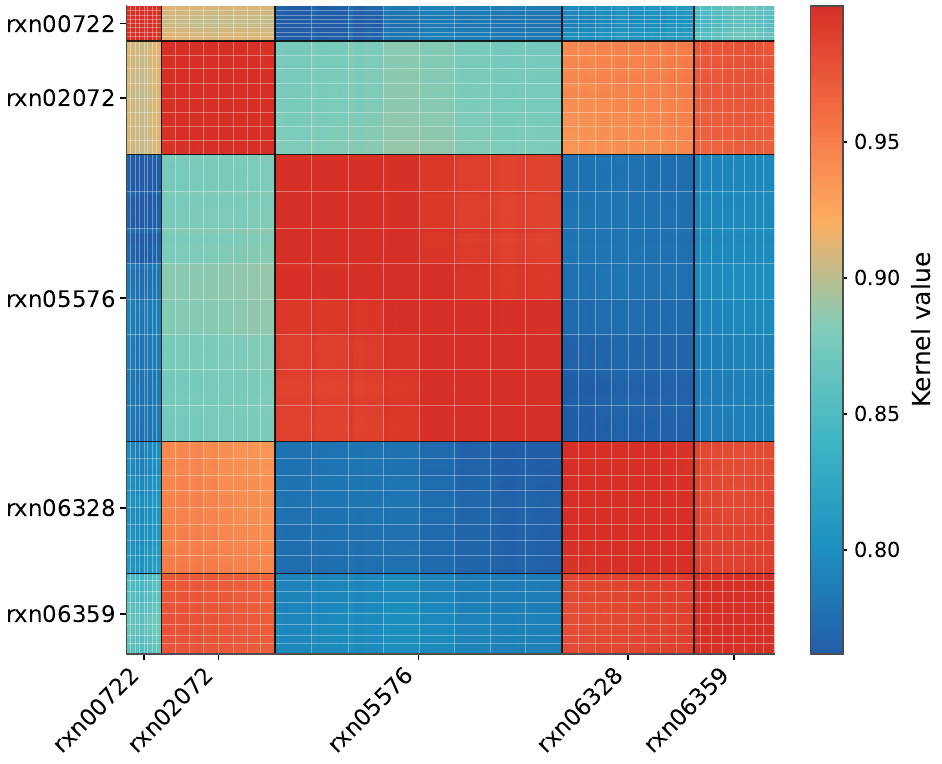}
      \caption{SOAP}
    \end{subfigure}
    \hfill
    \begin{subfigure}[t]{0.32\textwidth}
      \centering
      \includegraphics[width=\linewidth]{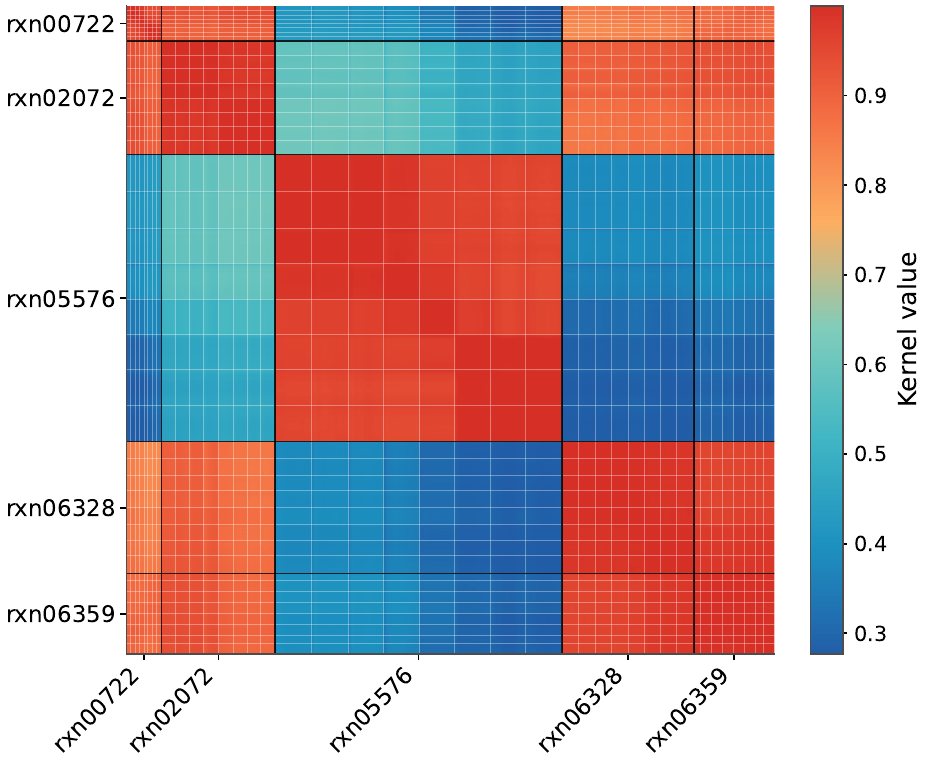}
      \caption{NTK, random init}
    \end{subfigure}
    \hfill
    \begin{subfigure}[t]{0.32\textwidth}
      \centering
      \includegraphics[width=\linewidth]{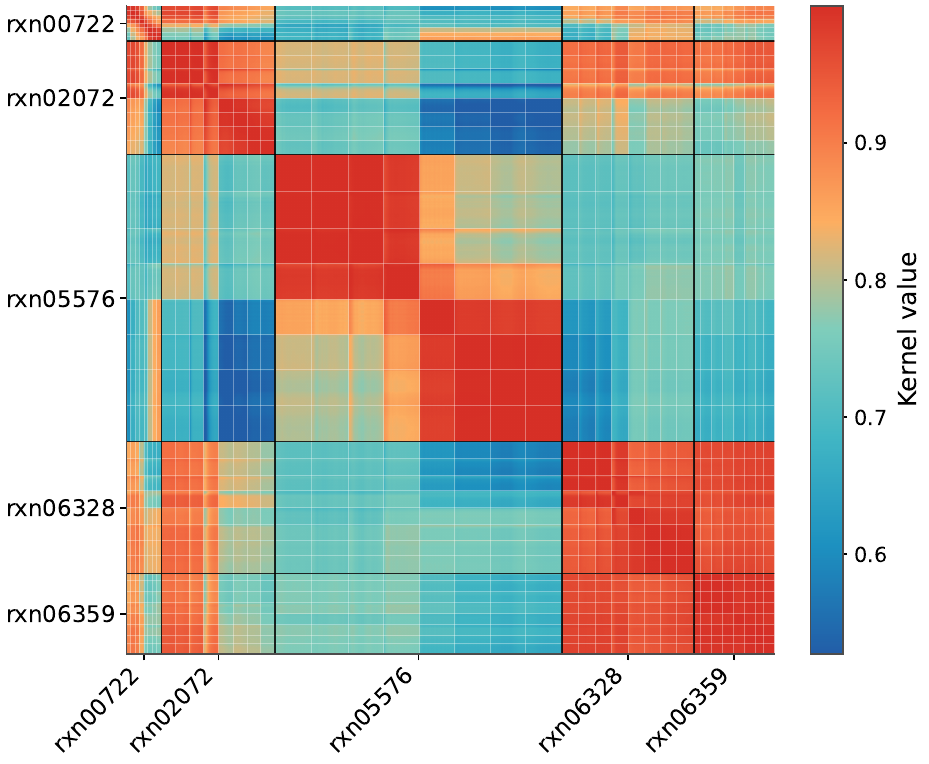}
      \caption{NTK, pretrained}
    \end{subfigure}
    \caption{Global kernel matrices on a five-reaction T1x subset. Structures are
    sorted by reaction family and frame index. The pretrained NTK preserves coarse
    reaction-family structure while retaining finer variation along reaction
    paths. Appendix~\ref{app:full-natural-bias} gives the full global and
    within-reaction kernel diagnostics for NTK, activations, SOAP, and Tanimoto kernels.}
    \label{fig:kernel-geometry}
  \end{figure*}

The acquisition results suggest that the pretrained model induces a useful
similarity space for reactive structures. Figure~\ref{fig:kernel-geometry}
visualises this geometry on a five-reaction T1x subset. Structures are ordered
first by reaction family and then by frame index along the reaction coordinate.
A useful kernel should capture both levels of structure: it should separate
different reaction families while also resolving meaningful variation within a
single reaction path.

Among the kernels visualised in Figure \ref{fig:kernel-geometry}, the pretrained NTK most clearly combines these two properties. It
preserves reaction-family block structure, but does not collapse each family into
a nearly homogeneous cluster; instead, it retains variation along the reaction
coordinate. The randomly initialised NTK shows some related architectural signal,
but is substantially more homogeneous, indicating that pretraining sharpens the
representation geometry. SOAP recovers coarse reaction-family structure, but is
comparatively saturated within individual reaction paths. Activation kernels show
similar behaviour to the NTK. Full global and within-reaction kernel
diagnostics for NTK, activations, SOAP, and Tanimoto are provided in
Appendix~\ref{app:full-natural-bias}.

\subsection{Residual-GP Uncertainty Calibration}
\label{sec:residual-gp-uncertainty-calib}
As another diagnostic, we test whether the kernels used for acquisition also provide useful residual uncertainty estimates. Well-calibrated uncertainty is important for kernel-based acquisition strategies: it ensures that predictive variances correspond to true error frequencies, which, in turn, enables principled decision-making. 

For each kernel, we hold the pretrained MACE prediction $b(\mathbf{x})$ fixed and fit a Gaussian process (GP) to the residual target $r(\mathbf{x}) = y - b(\mathbf{x})$. The GP therefore uses the kernel only to model the residual correction on top of the pretrained predictor. Because the kernel determines the geometry, but not the overall scale of the residual signal, we estimate a variance scale from the training residuals and evaluate Gaussian negative log-likelihood (NLL), predictive intervals, and calibration metrics using the resulting predictive variance. Details can be found in Appendix \ref{app:residual_calibration}.

We compare six residual kernels: pretrained activation features, randomly initialised activation features, pretrained NTK, randomly initialised NTK, SOAP, and Tanimoto. For each kernel, we use the Set 0 dataset from the T1x case study in Section \ref{sec: t1x_case_study} and replay posterior-variance acquisition from the same initial seed to obtain a nested sequence of training prefixes. The model-based kernels are computed once and held fixed; we do not fine-tune the model at each step. We do this to isolate the effects of pretraining on the model inductive bias, but it is also computationally cheaper since no training is required. 

\begin{figure}[h]
    \centering
    \includegraphics[width=0.8\linewidth]{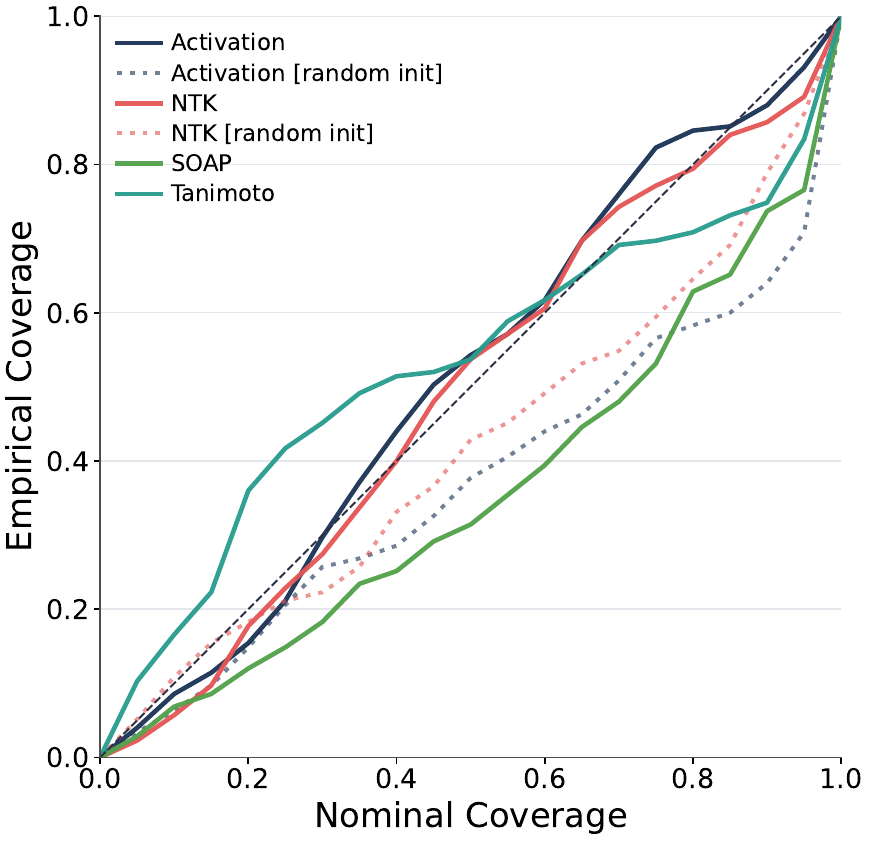}
\caption{Nominal versus empirical
coverage, for the selected kernels. Pretrained
activation and pretrained NTK show the best calibration among the compared kernels, while random neural kernels and descriptor kernels are less accurate
and less well calibrated.}
    \label{fig:calibration}
  \end{figure}

  \begin{table}[t]
    \centering
    \scriptsize
    \setlength{\tabcolsep}{4pt}
      \begin{tabular}{lccccc}
\toprule
Kernel & G. NLL & ECE & ENCE & 90\% Cov. & RMSE (meV) \\
\midrule
Activation            & \textbf{-1.679} & 0.028          & \textbf{0.237} & \textbf{0.880} & \textbf{45.44} \\
Activation (random)   & 0.137           & 0.106          & 0.796          & 0.640          & 155.83 \\
NTK                   & -1.439          & \textbf{0.020} & 0.337          & 0.857          & 54.54 \\
NTK (random init)     & -0.917          & 0.071          & 0.294          & 0.789          & 88.95 \\
SOAP                  & 0.837           & 0.113          & 0.741          & 0.737          & 351.90 \\
Tanimoto              & 2.741           & 0.069          & 1.573          & 0.749          & 253.86 \\
\bottomrule
\end{tabular}
\caption{Residual-GP test metrics at the validation-selected stopping point.
Lower is better for Gaussian NLL, RMSE, ECE, and ENCE, with ECE and ENCE equal
to zero indicating perfect calibration. Empirical 90\% coverage is best when
closest to the nominal 90\% level.}
    \label{tab:gp-calibration}
  \end{table}
  
Figure~\ref{fig:calibration} and Table~\ref{tab:gp-calibration} show that the pretrained model-based kernels provide the strongest residual uncertainty estimates. Pretrained activation features give the best Gaussian NLL, expected normalised calibration error (ENCE), 90\% coverage, and root mean squared error (RMSE), while the pretrained NTK gives the lowest expected calibration error (ECE). Randomly initialised neural kernels are weaker than their pretrained counterparts, and descriptor kernels are weaker overall. These results are consistent with the mechanism suggested by the acquisition experiments: pretraining induces representation geometries that are better aligned with the residual errors left by the MACE bias model.

Note that because the GP is fit with a global structure kernel, its residual predictions should not be interpreted as a fully accurate energy model: the true energy depends on local atomic environments. We expect this limitation to become more pronounced for more difficult fine-tuning tasks beyond T1x. In such settings, the residual function may change substantially during fine-tuning. Thus, the residual GP should be interpreted as a diagnostic of how well the pretrained representation organises residual error, rather than as a fully adaptive uncertainty model for the fine-tuned MLIP.

\subsection{Additional Reactivity Datasets}
\label{sec:transferability}

In this section we test whether model-based acquisition strategies outperform random acquisition across other reactivity benchmarks. These transfer experiments keep the same pretrained model family and therefore test dataset transfer rather than architecture transfer.

We evaluate the same pretrained model on PMechDB~\cite{Tavakoli2024},
RGD~\cite{Zhao2023}, and a larger T1x subset. RGD and PMechDB are random 5k and 10k splits
of the corresponding OMol subsets~\cite{levine2026openmolecules2025omol25}, with PMechDB filtered to H, C, N, and O chemistry so we can use the pretrained SPICE 2 model. The larger T1x subset,
which we call \textit{T1x Mixed}, contains a random selection of 100 T1x reaction pathways.

For each dataset, we first reserve disjoint test and validation sets comprising \(20\%\) and \(10\%\) of the data, respectively. The remaining structures define the acquisition pool. From this pool, we initialise active learning with 50 labelled structures and then run 20 acquisition rounds, adding 150 structures per round. Each experiment is repeated with two random seeds and metrics are averaged over both seeds. We compare model-based acquisition families to committee and random-based acquisition. For kernel-based methods, we use LCMD because posterior-variance selection can
over-prioritise isolated high-variance outliers, whereas LCMD balances distance
from the labelled set with cluster mass, and is more stable on heterogeneous candidate pools. 

Table~\ref{tab:transfer_benchmarks} reports round gains relative to random acquisition. For each dataset,
method, and metric, we define the best shared value as the lowest value reached by both the method and random, and we report the percentage gain in the number of rounds taken to reach this target. Positive values indicate that the method reaches the target earlier than random,
whereas negative values indicate that it reaches the target later.

The model-based representation methods transfer most consistently, whereas the energy committee shows mixed behaviour across energy and force metrics. Specifically, both Activation-LCMD and NTK-LCMD improve force RMSE on all three datasets, and they also improve energy RMSE and energy MAE across all three benchmarks. They achieve the strongest average gains across metrics overall. The gains are more modest on RGD and PMechDB than on T1x, which we believe reflects the greater difficulty of those settings. The gains are also larger for RMSE than for MAE, consistent with active learning preferentially reducing high-error tail cases. Detailed learning curves are provided in Appendix~\ref{sec:appendix-transferability}.

  \begin{table}[t]
      \centering
      \scriptsize

\definecolor{posgreen}{RGB}{34,139,34}
\definecolor{negred}{RGB}{180,50,47}

\newcommand{\pos}[1]{\textcolor{posgreen}{#1}}
\newcommand{\negv}[1]{\textcolor{negred}{#1}}

\setlength{\tabcolsep}{2pt}
\begin{tabular}{llcccc}
\toprule
\textbf{Method} & \textbf{Metric} & \textbf{PMechDB} & \textbf{RGD} & \textbf{T1X Mixed} & \textbf{Average} \\
\midrule

\multirow{4}{*}{\textbf{Activation LCMD}}
& E RMSE & \pos{+15.0\%} & \pos{+22.2\%} & \pos{+77.8\%} & \textbf{\pos{+38.3\%}} \\
& F RMSE & \pos{+20.0\%} & \pos{+10.0\%} & \pos{+44.4\%} & \textbf{\pos{+24.8\%}} \\
& E MAE  & \pos{+5.0\%}  & \pos{+10.0\%} & \pos{+61.5\%} & \textbf{\pos{+25.5\%}} \\
& F MAE  & +0.0\%        & +0.0\%        & \pos{+15.0\%} & \textbf{\pos{+5.0\%}} \\
\midrule

\multirow{4}{*}{\textbf{NTK LCMD}}
& E RMSE & \pos{+20.0\%} & \pos{+11.1\%} & \pos{+83.3\%} & \textbf{\pos{+38.1\%}} \\
& F RMSE & \pos{+25.0\%} & \pos{+10.0\%} & \pos{+50.0\%} & \textbf{\pos{+28.3\%}} \\
& E MAE  & \pos{+5.0\%}  & \pos{+15.0\%} & \pos{+61.5\%} & \textbf{\pos{+27.2\%}} \\
& F MAE  & +0.0\%        & +0.0\%        & \pos{+25.0\%} & \textbf{\pos{+8.3\%}} \\
\midrule

\multirow{4}{*}{\textbf{Committee Energy}}
& E RMSE & \negv{-23.5\%} & \pos{+11.1\%} & \pos{+27.8\%} & \textbf{\pos{+5.1\%}} \\
& F RMSE & \pos{+25.0\%}  & \pos{+15.0\%} & +0.0\%        & \textbf{\pos{+13.3\%}} \\
& E MAE  & \negv{-31.6\%} & \pos{+5.0\%}  & \negv{-13.3\%} & \textbf{\negv{-13.3\%}} \\
& F MAE  & \negv{-5.0\%}  & +0.0\%        & +0.0\%        & \textbf{\negv{-1.7\%}} \\
\bottomrule
\end{tabular}

\caption{Round gain relative to random acquisition across datasets. Positive values indicate
earlier acquisition than random, and negative values indicate later acquisition.}
  \label{tab:transfer_benchmarks}
  \end{table}

\section{Discussion}

Across the reactive-chemistry settings we study, kernels built directly from pretrained or fine-tuned MACE models, that is the energy NTK and hidden activations, give stronger active-learning performance than fixed molecular descriptors or committee disagreement based strategies. Kernel diagnostics support the same interpretation: pretrained neural representations preserve both reaction-family structure and within-reaction variation, the distinctions needed to select informative structures along reaction pathways. Concretely, this means that model-derived representations are robust under the kind of generator-induced pool bias that reactive chemistry intrinsically exhibits.

From a practical perspective, activation-based representations are especially attractive as they require only a forward pass, whereas NTK features require an additional backward pass but remain substantially cheaper than committee ensembles, which add $M$ full training runs per round. Both representations are already present inside the model being fine-tuned and require no auxiliary training.

The binding constraint of our implementation is memory. Both PV and LCMD operate in kernel space and form the candidate--candidate Gram matrix $k_{\mathcal{P}\mathcal{P}}\!\in\!\mathbb{R}^{n_\mathcal{P}\times n_\mathcal{P}}$, which costs $O(n_\mathcal{P}^2)$. At the moderate pool sizes considered here ($n_\mathcal{P}\!\le\!10$k), $k_{\mathcal{P}\mathcal{P}}$ comfortably fits on a single GPU; an order of magnitude more candidates and the kernel can no longer be materialised on a single device, making kernel construction the dominant cost. Activation features are cheaper per byte ($d{=}256$, $\sim\!10$\,MB at $n_\mathcal{P}{=}10$k) but the kernel itself remains $O(n_\mathcal{P}^2)$, so they share the same large-pool ceiling.

Future work should aim to extend to larger candidate pools, and test whether these conclusions extend to other pretrained MLIP architectures and datasets, and consider alternative committee constructions in pretrained settings. It would also be valuable to study force-aware model representations, compare against newer uncertainty heads~\cite{ho2025flexibleuncertaintycalibrationmachinelearned}, Bayesian interatomic potentials~\cite{coscia2026blipsbayesianlearnedinteratomic}, and multi-head ensemble methods~\cite{beck2025multihead}.

More broadly, the same kernels could be useful beyond active learning, for training-set summarisation, dataset distillation and validation-set construction. Across these settings the underlying question is the same: whether pretrained molecular representations define a similarity space better aligned with downstream error than fixed descriptors alone. Our results suggest this direction is promising.

\section*{Acknowledgements}
We thank Silvia Acosta Gutiérrez, Massimo Bortone, Heloise Chomet, Valentin Heyraud, Jack Simons, Tamás Lajos Tompa, Lucien Walewski, and Leon Wehrhan for insightful discussions. We also thank Scott Cameron for highlighting the potential usefulness of Neural Tangent Kernels for uncertainty quantification.
\bibliographystyle{icml2026}
\bibliography{main}

\clearpage
\onecolumn
\appendix

\section{MACE Architecture and Representation Extraction}
\label{app:mace-architecture}

We write a structure as $\mathbf{x}=(\mathbf{z},\mathbf{r})$, with atomic
numbers $\mathbf{z}=(z_i)_{i=1}^{N(\mathbf{x})}$ and Cartesian coordinates
$\mathbf{r}=(\mathbf{r}_i)_{i=1}^{N(\mathbf{x})}$. MACE \cite{batatia2023macehigherorderequivariant} constructs a
radius-cutoff graph $G(\mathbf{x})$: nodes are atoms, and directed edges connect
neighbours within cutoff $r_{\max}$. Node attributes are one-hot species vectors
$\mathbf{a}_i$. For an edge $j\to i$, define the relative vector
$\mathbf{r}_{ji}=\mathbf{r}_j-\mathbf{r}_i$, distance
$d_{ji}=\|\mathbf{r}_{ji}\|$, and direction
$\hat{\mathbf{r}}_{ji}=\mathbf{r}_{ji}/d_{ji}$.

The embedding stage has two parts: an atomic (node) embedding block and a
radial (edge) embedding block.

\paragraph{Atomic embedding weights.}
The atomic embedding block maps the one-hot species attribute $\mathbf{a}_i$ of
each atom to an initial scalar node feature. Concretely, it is a learnable
lookup table
\begin{equation}
  W^{\mathrm{emb}} \in \mathbb{R}^{Z_{\max}\times c},
  \qquad
  h_{i,L=0}^{(0)} \;=\; W^{\mathrm{emb}}_{z_i,\,:},
\end{equation}
where $Z_{\max}$ is the number of supported atomic species and $c$ is the
number of scalar channels. The initial node feature of atom $i$ is therefore
the row of $W^{\mathrm{emb}}$ indexed by its atomic number $z_i$: it depends
only on the atomic species, not on the local environment, the geometry, or
the rest of the structure. Two atoms of the same species in different
configurations enter the network with identical initial features, and
differences in their downstream representations arise entirely from the
geometric information injected by the interaction blocks. Equivalently, the
embedding weights act as a per-species learnable bias that carries the
chemical identity of the atom into the rest of the model.

\paragraph{Radial embedding.}
The radial embedding block maps each distance $d_{ji}$ to edge features
through a Bessel basis modulated by a smooth cutoff envelope. In the MACE
checkpoints used in this work this basis is parameter-free, so the trainable
parameters of the embedding stage are dominated by $W^{\mathrm{emb}}$.
Angular information is injected separately, and not at the embedding stage,
through spherical harmonics of $\hat{\mathbf{r}}_{ji}$.

\paragraph{Interaction stack.}
Interaction blocks combine neighbour node features with radial and angular
edge features to produce equivariant messages, and product blocks increase
the effective body order through symmetric tensor products. After interaction layer
$\ell$, the hidden state of atom $i$ can be written as
\begin{equation}
  h_i^{(\ell)}
  =
  \bigoplus_{L=0}^{L_{\max}} h_{i,L}^{(\ell)},
\end{equation}
where $L=0$ channels are invariant scalars and $L>0$ channels transform as
vectors or higher-order tensors under rotations.

MACE predicts energy through scalar readouts applied to the invariant channels.
With $T$ interaction layers, the total energy has the form
\begin{equation}
  E_\theta(\mathbf{x})
  =
  \sum_{i=1}^{N(\mathbf{x})}
  \sum_{\ell=0}^{T}
  E_i^{(\ell)}(\mathbf{x}),
\end{equation}
and forces are analytic coordinate gradients,
\begin{equation}
  \mathbf{F}_i(\mathbf{x}) = -\nabla_{\mathbf{r}_i} E_\theta(\mathbf{x}).
\end{equation}
The activation representation in the main text pools the scalar
$h_{i,L=0}^{(\ell)}$ channels over atoms and concatenates the pooled features
from each message-passing layer. In our experiments, the MACE model uses angular features up to $L=2$ and has two
message-passing layers with 128 scalar hidden channels per layer, so this gives
a 256-dimensional activation vector. The energy NTK representation differentiates
$E_\theta(\mathbf{x})$ with respect to the node embedding
parameters, which are the same MACE blocks that encode species identity. Although these parameters are themselves
species-indexed and geometry-free, the corresponding gradient is computed by
backpropagating the energy through the full interaction stack, so each row of
$\nabla_{W^{\mathrm{emb}}} E_\theta(\mathbf{x})$ aggregates per-atom
sensitivities that depend on the entire local environment of every atom of that species. 
The embedding-NTK feature can therefore distinguish structures with identical composition but different geometries, while remaining substantially smaller than the gradient with respect to the full network.

\section{Training Details}
\label{app:training-details}

All active-learning rounds fine-tune the same SPICE 2 pretrained MACE checkpoint
that is trained using the \texttt{mlip} library~\cite{brunken2025machinelearninginteratomicpotentials}.
Weights are optimised on the current labelled set without freezing any model
parameters. The surrogate predicts total energies and atomic forces, and the
training objective is a Huber loss on energy and force errors with equal weights
$w_E=w_F=10.0$.

For the T1x dataset, each round uses a dynamic schedule that keeps the
approximate number of gradient updates comparable as the labelled set grows. If
$|\mathcal{T}^{(t)}|$ is the current labelled-set size, the batch size $B$ and
learning rate $\eta$ are
\begin{equation}
    (B,\eta)=
\begin{cases}
(1,10^{-3}), & |\mathcal{T}^{(t)}|\le 20,\\
(2,5\times 10^{-3}), & 20<|\mathcal{T}^{(t)}|\le 100,\\
(4,5\times 10^{-3}), & |\mathcal{T}^{(t)}|>100,
\end{cases}
\end{equation}

and the number of epochs is
\begin{equation}
    E=\max\!\left(10,\left\lceil\frac{1000B}{|\mathcal{T}^{(t)}|}\right\rceil\right).
\end{equation}

For all other datasets, we train for 50 epochs with a batch size of 16 and learning rate of $0.01$. This was sufficient to get approximate convergence for all trainings.

The initial training sets, validation sets, test sets, and candidate pool are
disjoint. Validation performance is used for model selection at each
active-learning round.

\section{Full T1x results}
\label{app:full-natural-bias}

Table~\ref{tab:natural_bias_grouped_kernel_method} reports the full T1x experimental results, including posterior
variance (PV), largest-cluster maximum-distance (LCMD), and k-center
(Kc)~\cite{Vazirani2003} variants. This is the full table underlying the abbreviated main-text summary
in Figure~\ref{fig:natural-t1x}.

\begin{table}[h]
\centering
\setlength{\tabcolsep}{8pt}
\caption{T1x summary metrics grouped by kernel family and method
variant. Final columns are final-round RMSE
in meV for energy and meV\,\AA$^{-1}$ for forces. AUC is the discrete sum of
RMSE over acquisition steps, with units of the corresponding RMSE. Best (lowest) values in each metric are shown in bold. The
reported values are averaged over the 3 sets.}
\label{tab:natural_bias_grouped_kernel_method}
\begin{tabular}{llcccc}
\toprule
Kernel & Method & Energy AUC ($\downarrow$) & Force AUC ($\downarrow$) & Final E RMSE ($\downarrow$) & Final F RMSE ($\downarrow$) \\
\midrule

Random     & --   & 648.33          & 6490.00          & 6.33          & 111.83          \\

\addlinespace[2pt]

Committee  & Energy & 623.33          & 6636.67          & 3.50          & 82.50           \\
           & Force  & 951.33          & 5496.00          & 8.00          & 67.67           \\

\addlinespace[2pt]

Activation & Kc    & 259.67          & 3208.67          & 3.00          & 43.83           \\
           & LCMD  & \textbf{252.00} & 3470.67          & \textbf{2.83} & 47.33           \\
           & PV    & 252.17          & 3179.00          & 3.17          & 42.67           \\

\addlinespace[2pt]

NTK        & Kc    & 273.50          & 3341.17          & \textbf{2.83} & 45.33           \\
           & LCMD  & 259.17          & 3452.00          & \textbf{2.83} & 45.67           \\
           & PV    & 252.83          & \textbf{3078.00} & \textbf{2.83} & \textbf{40.00}  \\

\addlinespace[2pt]

SOAP       & Kc    & 273.17          & 4080.83          & 3.50          & 54.00           \\
           & LCMD  & 278.83          & 4270.67          & 3.00          & 57.67           \\
           & PV    & 331.83          & 6926.17          & 4.33          & 126.50          \\

\addlinespace[2pt]

Tanimoto   & Kc    & 311.17          & 4580.00          & 4.00          & 84.17           \\
           & LCMD  & 707.67          & 5209.50          & 5.67          & 78.17           \\
           & PV    & 275.33          & 3791.83          & 3.17          & 61.17           \\

\bottomrule
\end{tabular}
\end{table}
Table \ref{tab:cost_comparison} provides a practical cost comparison of the acquisition methods, including the representation dimension, the number of forward and backward passes, the number of required models, and the overall evaluation cost in the pretrained setting.

\begin{table}[h]
\centering
\caption{
Operational cost comparison of the acquisition families used in this work. For each method we report the number of forward and backward passes per candidate, the number of models maintained, the representation dimension $d$, the binding peak-memory term in our kernel-space implementation, and an overall practical cost category. Kernel methods (PV, LCMD) form an $n_\mathcal{P}\!\times\!n_\mathcal{P}$ Gram matrix; this is the dominant memory cost and scales as $O(n_\mathcal{P}^2)$ regardless of $d$. Committees instead pay $M$ full training runs per round.
}
\label{tab:cost_comparison}
\setlength{\tabcolsep}{3pt}
\begin{tabular}{lccclll}
\toprule
Method & Fwd. & Bwd. & \#Models & Dim. $d$ & Peak memory & Practical cost \\
\midrule
Activation     & 1   & 0 & 1   & 256   & $O(n_\mathcal{P}^2)$ kernel & Low--Moderate \\
NTK            & 1   & 1 & 1   & 1{,}920 & $O(n_\mathcal{P}^2)$ kernel + $O(n_\mathcal{P} d)$ feats & Moderate \\
SOAP           & 0   & 0 & 1   & 3{,}696 & $O(n_\mathcal{P}^2)$ kernel & Low \\
Tanimoto       & 0   & 0 &1   & 2{,}048 (binary) & $O(n_\mathcal{P}^2)$ kernel & Low \\
Committee-E    & $M$ & 0 & $M$ & --    & $M{\times}$ model state    & High ($M$ trainings/round) \\
Committee-F    & $M$ & 0 & $M$ & --    & $M{\times}$ model state    & High ($M$ trainings/round) \\
Random         & 0   & 0 & 1   & --    & --                         & Minimal \\
\bottomrule
\end{tabular}
\end{table}

\subsection{Committee benchmarks}
\label{sec:appendix-committee-natural-bias}

We benchmark committee acquisition under the T1x setting using the same active-learning splits and training schedule as the other methods. Each
committee contains three MACE models initialised from the same pretrained checkpoint and trained on the current labelled set at each acquisition round.

We compare two ways of inducing diversity across committee members. In the
\textit{shuffle} variant, all committee members see the same labelled set but
use independent data-order seeds during training. In the \textit{bootstrap}
variant, each member is trained on a sample drawn with replacement from the
current labelled set, so individual ensemble members see different empirical
training distributions. Both variants are evaluated with energy and force
disagreement scores.

\begin{figure}[h]
    \centering
    \includegraphics[width=\linewidth]{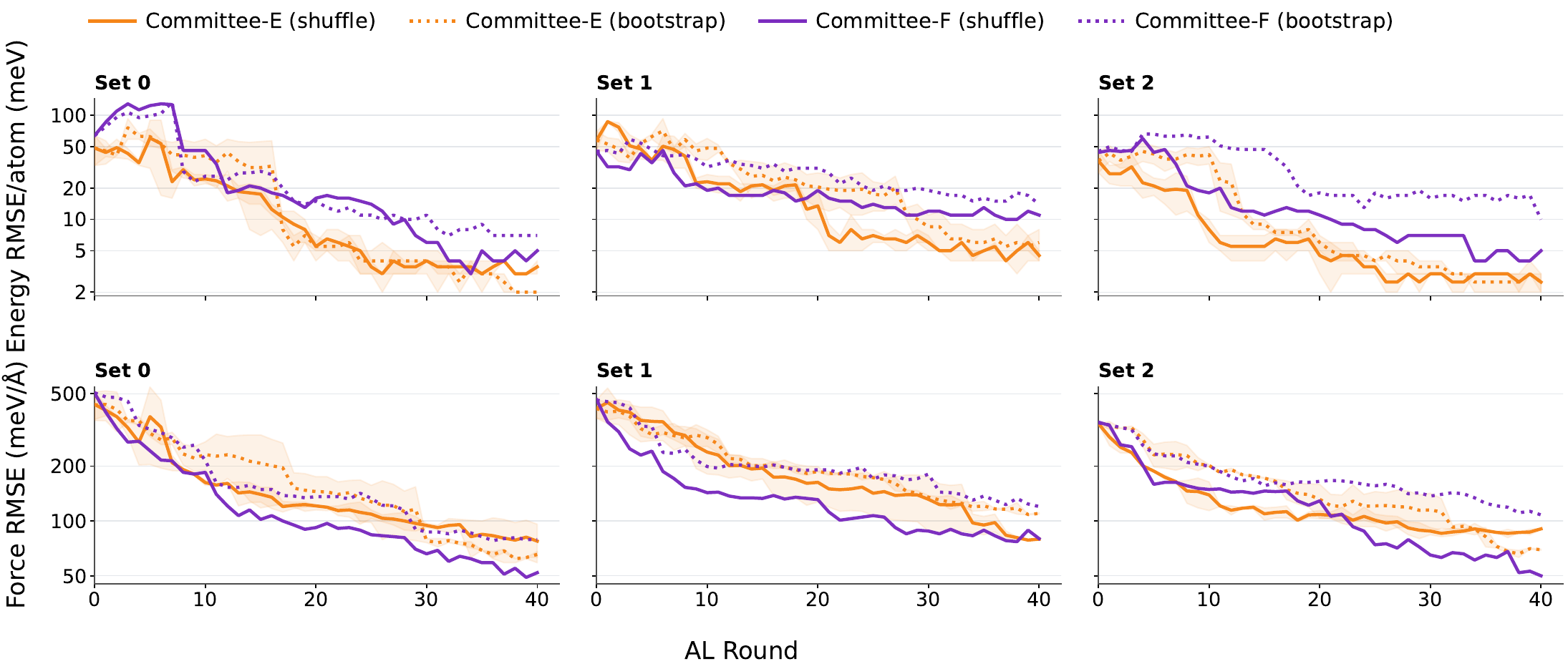}
    \caption{Energy and force RMSE across active-learning rounds for committee
    variants on the three T1x sets. We compare energy and force
    disagreement scores, each with shuffle- and bootstrap-based ensemble
    diversity, across the three natural-bias candidate pools.}
    \label{fig:committee-learning-curves}
\end{figure}

Figure~\ref{fig:committee-learning-curves} shows the energy and force learning
curves for these committee variants on the three T1x candidate pools.
The main failure mode is an unstable energy--force trade-off. Energy
committees can improve final energy error, but their acquisition scores are not
well aligned with force improvement and they give weaker force learning curves
than random acquisition. Force committees select
structures that are more useful for reducing force error, but this comes at a
large cost to energy accuracy: in
Table~\ref{tab:natural_bias_grouped_kernel_method}, Committee-F has better
final force error than Committee-E, but has the worst energy AUC and final
energy error among the reported natural-bias methods.

\begin{figure}[h]
    \centering
    \includegraphics[width=\linewidth]{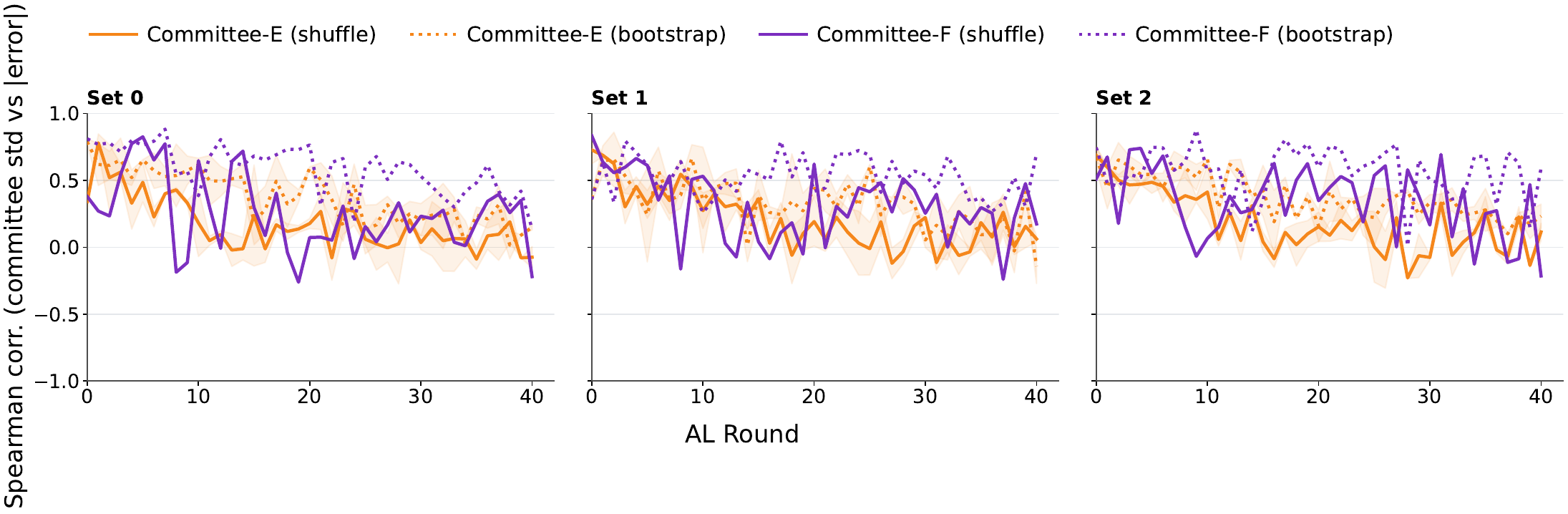}
    \caption{Spearman correlation between committee standard deviation and
    absolute prediction error across active-learning rounds. The weak correlations show that committee disagreement is not a consistently
    calibrated ranking signal.}
    \label{fig:committee-spearman}
\end{figure}

The rank-correlation diagnostic in
Figure~\ref{fig:committee-spearman} helps explain this behaviour. At each
round, we compute the Spearman correlation between the committee standard
deviation on pool candidates and the corresponding absolute prediction error.
A useful uncertainty score should have a consistently positive correlation with
held-out error. Instead, the committee correlations are weak across
candidate pools, rounds, and randomization schemes. This indicates that the
committee score is often ranking candidates by ensemble variability that does
not correspond to the downstream error being optimized. 

\clearpage

\subsection{Scratch training}\label{sec: scratch}

In this section, we train models from scratch using the same MACE architecture with random initialization. For committee, this now involves using canonical MACE ensembles with different intialization seeds. Figure~\ref{fig:scratch_curves} shows the resulting learning curves. We observe qualitatively similar behavior to the pretrained setting: model-based acquisition strategies such as NTK and activations significantly outperform the alternatives. The full averaged results can be found in Table~\ref{tab:scratch_vs_pretrained_grouped}, which also indicates consistent improvements when using pretrained models compared to training from scratch.

\begin{center}
  \centering
  \includegraphics[width=\linewidth]{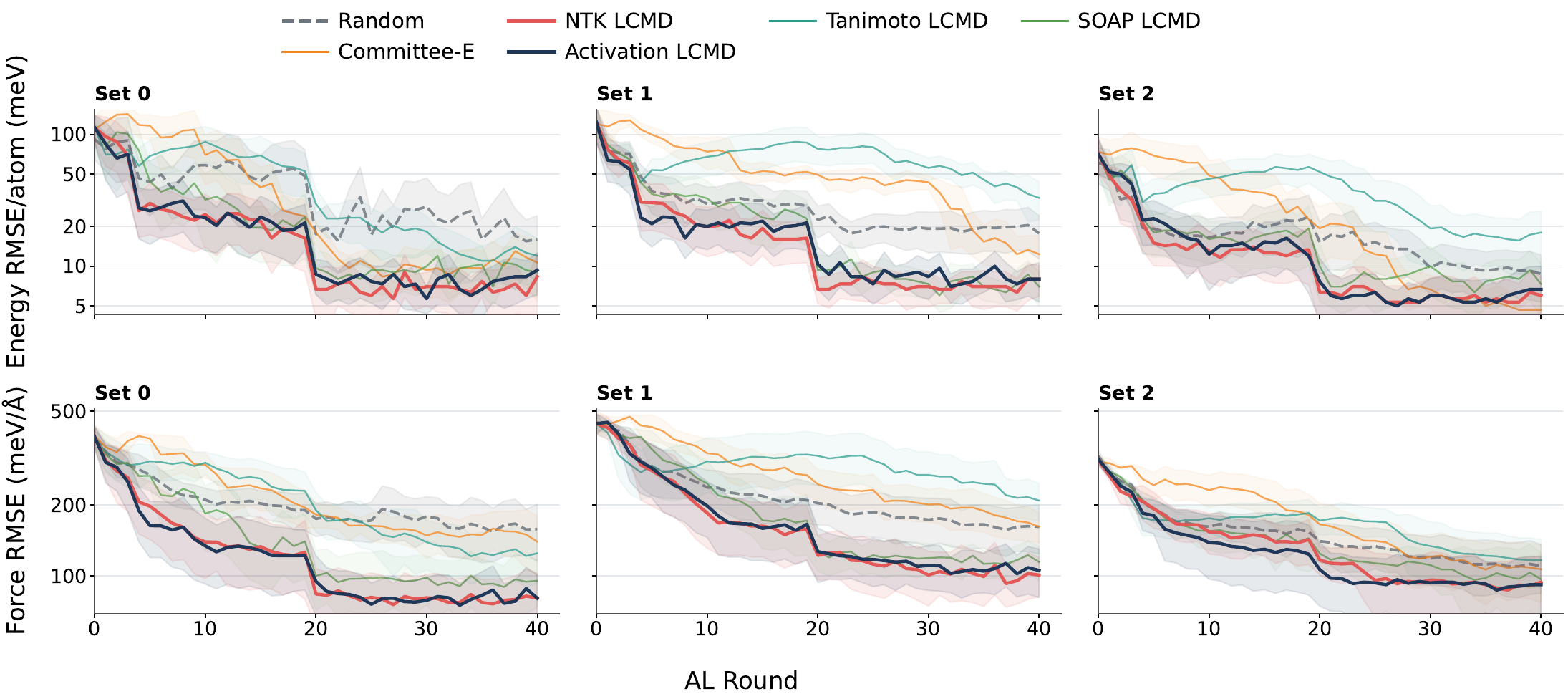}
  \captionof{figure}{Force and energy learning curves across the T1x sets when training from scratch showing similar results to the pretrained case.}
  \label{fig:scratch_curves}
\end{center}

\begin{table}[h]
\centering
\setlength{\tabcolsep}{8pt}
\caption{Comparison of scratch and pretrained performance grouped by kernel family. All kernel methods are run with LCMD acquisition. Final columns are RMSE in meV for energy and meV\,\AA$^{-1}$ for forces. Best (lowest) values in each metric are shown in bold. The reported values are averaged over the 3 sets.}
\label{tab:scratch_vs_pretrained_grouped}
\begin{tabular}{llcccc}
\toprule
Kernel & Method & Energy AUC ($\downarrow$) & Force AUC ($\downarrow$) & Final E RMSE ($\downarrow$) & Final F RMSE ($\downarrow$) \\
\midrule

Random & Scratch    & 1211.58         & 7953.25         & 14.17         & 143.17         \\
       & Pretrained & 648.33          & 6490.00         & 6.33          & 111.83         \\

\addlinespace[2pt]

Committee & Scratch    & 1803.44         & 9352.44         & 9.22          & 135.89         \\
          & Pretrained & 623.33          & 6636.67         & 3.50          & 82.50          \\

\addlinespace[2pt]

NTK & Scratch    & 743.22          & 5972.56         & 7.44          & 91.67          \\
    & Pretrained & 259.17          & \textbf{3452.00}        & \textbf{2.83} & \textbf{45.67}          \\

\addlinespace[2pt]

Activation & Scratch    & 764.22          & 5896.11         & 8.00          & 92.44          \\
           & Pretrained & \textbf{252.00} & 3470.67         & \textbf{2.83} & 47.33          \\

\addlinespace[2pt]

SOAP & Scratch    & 927.56          & 6724.44         & 7.78          & 102.11         \\
     & Pretrained & 278.83          & 4270.67         & 3.00          & 57.67          \\

\addlinespace[2pt]

Tanimoto & Scratch    & 2004.11         & 9201.11         & 21.00         & 149.89         \\
         & Pretrained & 707.67          & 5209.50         & 5.67          & 78.17          \\

\bottomrule
\end{tabular}
\end{table}
\clearpage

\raggedbottom
\subsection{Kernel visualisations}
We now present the detailed kernel visualisations underlying Figure~\ref{fig:kernel-geometry}. The global plots order candidate structures by reaction family, making it possible to assess whether a representation separates different reaction classes. The frame-block plots isolate within-reaction structure by ordering frames along each reaction coordinate. Together, these views distinguish coarse reaction-family separation from sensitivity to geometric changes along a reaction path.

The global kernel matrices are computed on the five-reaction T1x subset set 0. For both activations and NTK kernels, we compare trained models to their randomly initialised counterparts, and also track their evolution over selected fine-tuning iterations (1, 20, and 40).

The within-reaction frame-block plots isolate structure by displaying kernels restricted to individual reactions. These typically evolve during training. For example, for the NTK kernels shown in Figure~\ref{fig:ntk-frame-block-comparison}, iteration 1 shows some within-reaction variation across most reactions. By iterations 20 and 40, several reactions, most notably 00722, 05576, and 06328, develop clearer block patterns, where early frames become measurably less similar to later frames. In contrast, reactions 02072 and 06359 remain comparatively flat throughout, indicating weaker sensitivity to progression along the reaction coordinate.

\textbf{Pretrained NTK Kernels}

\begin{figure}[!htbp]
  \centering
  \includegraphics[width=\linewidth]{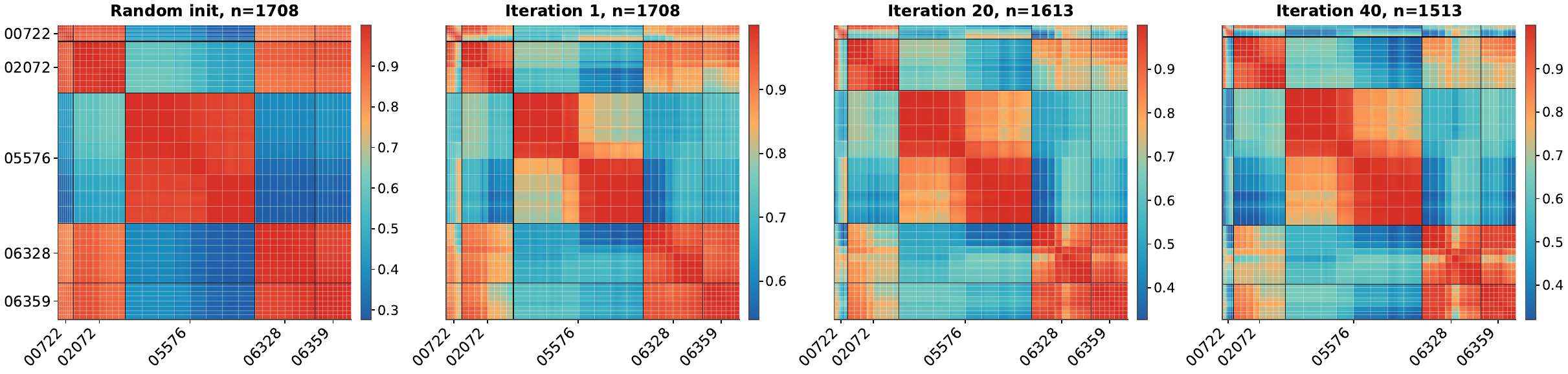}
  \caption{Global NTK kernel matrices on the T1x subset, with structures
  ordered by reaction family. The panels compare the random-initialised MACE NTK
  with NTK kernels computed after selected active-learning iterations.}
  \label{fig:ntk-global-comparison}
\end{figure}
\FloatBarrier

\begin{center}
  \centering
  \includegraphics[width=\linewidth]{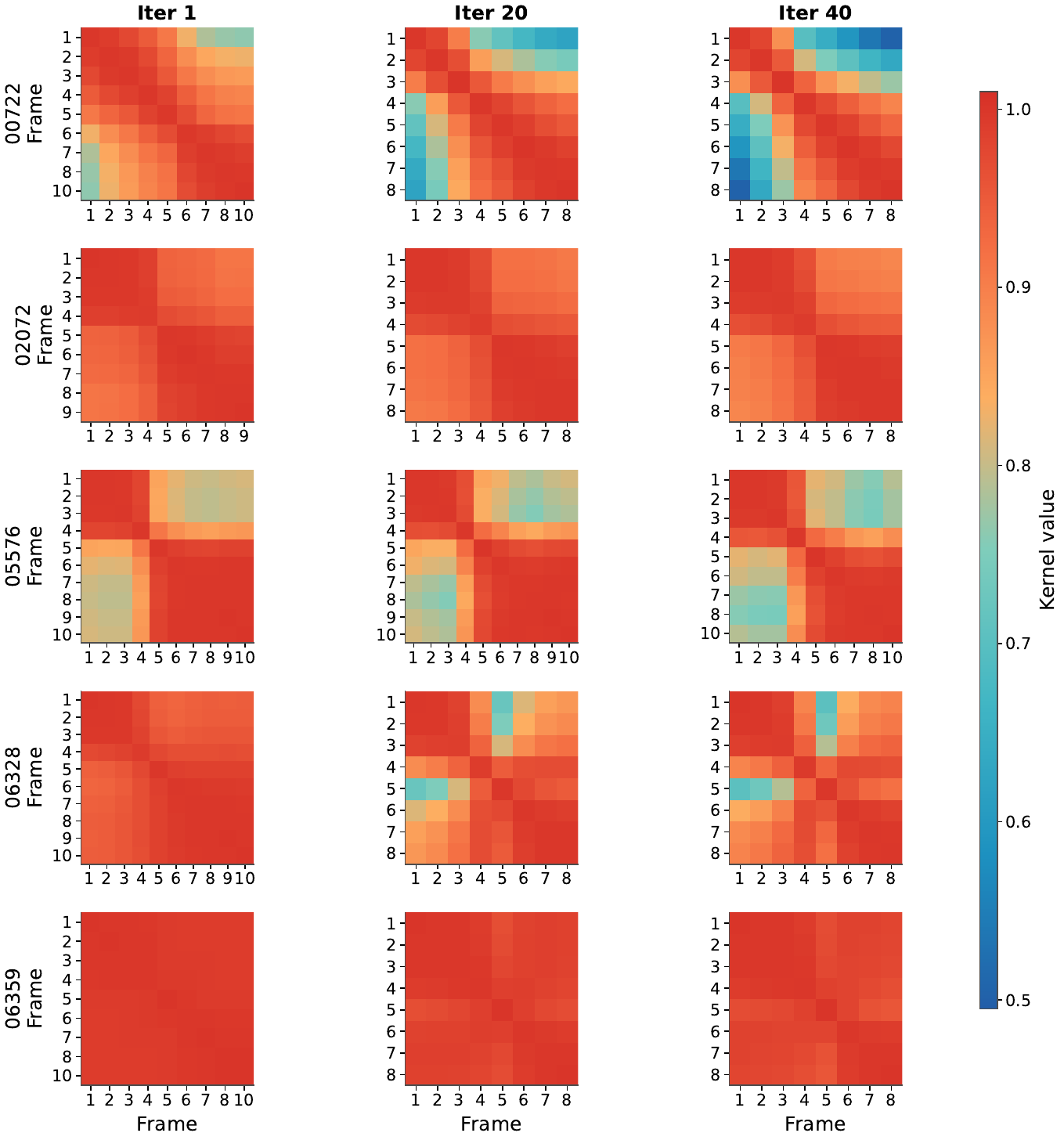}
  \captionof{figure}{Within-reaction NTK frame-block kernels for the T1x subset after
  selected active-learning iterations. Each block orders structures by frame
  index along a reaction pathway.}
  \label{fig:ntk-frame-block-comparison}
\end{center}

\clearpage

\textbf{Scratch NTK kernels}

\begin{center}
  \centering
  \includegraphics[width=\linewidth]{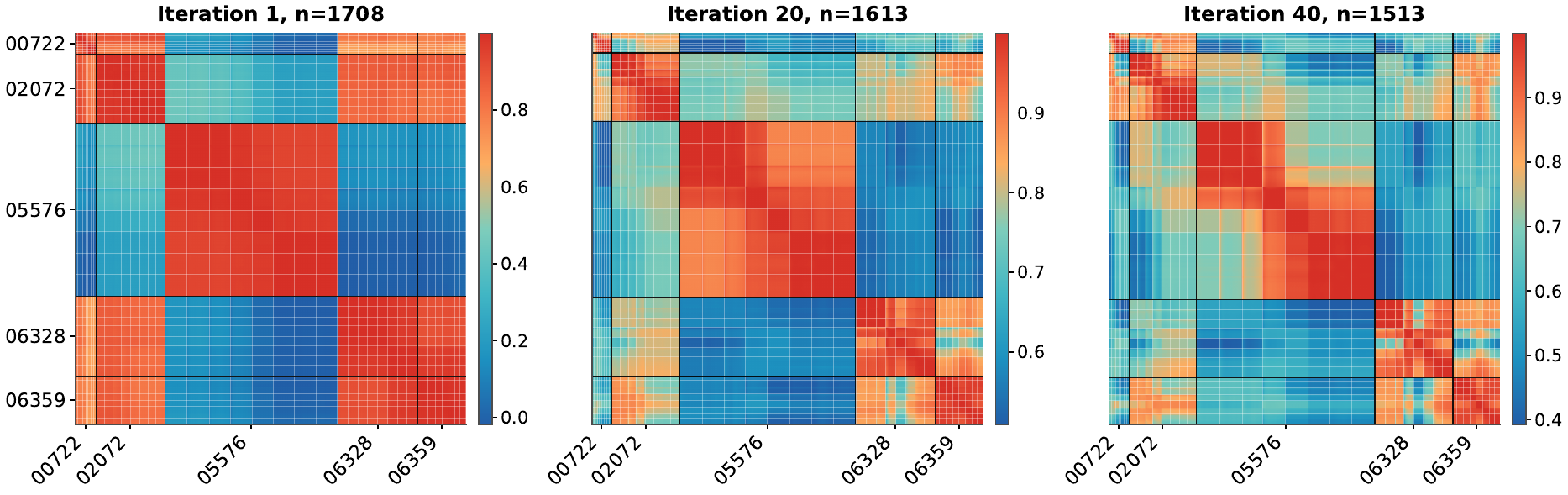}
  \captionof{figure}{Global NTK kernel matrices on the T1x subset, with structures
  ordered by reaction family. The panels show how an untrained NTK evolves with AL iteration.}
  \label{fig:scratchntk-global-comparison}
\end{center}

\begin{center}
  \centering
  \includegraphics[width=\linewidth]{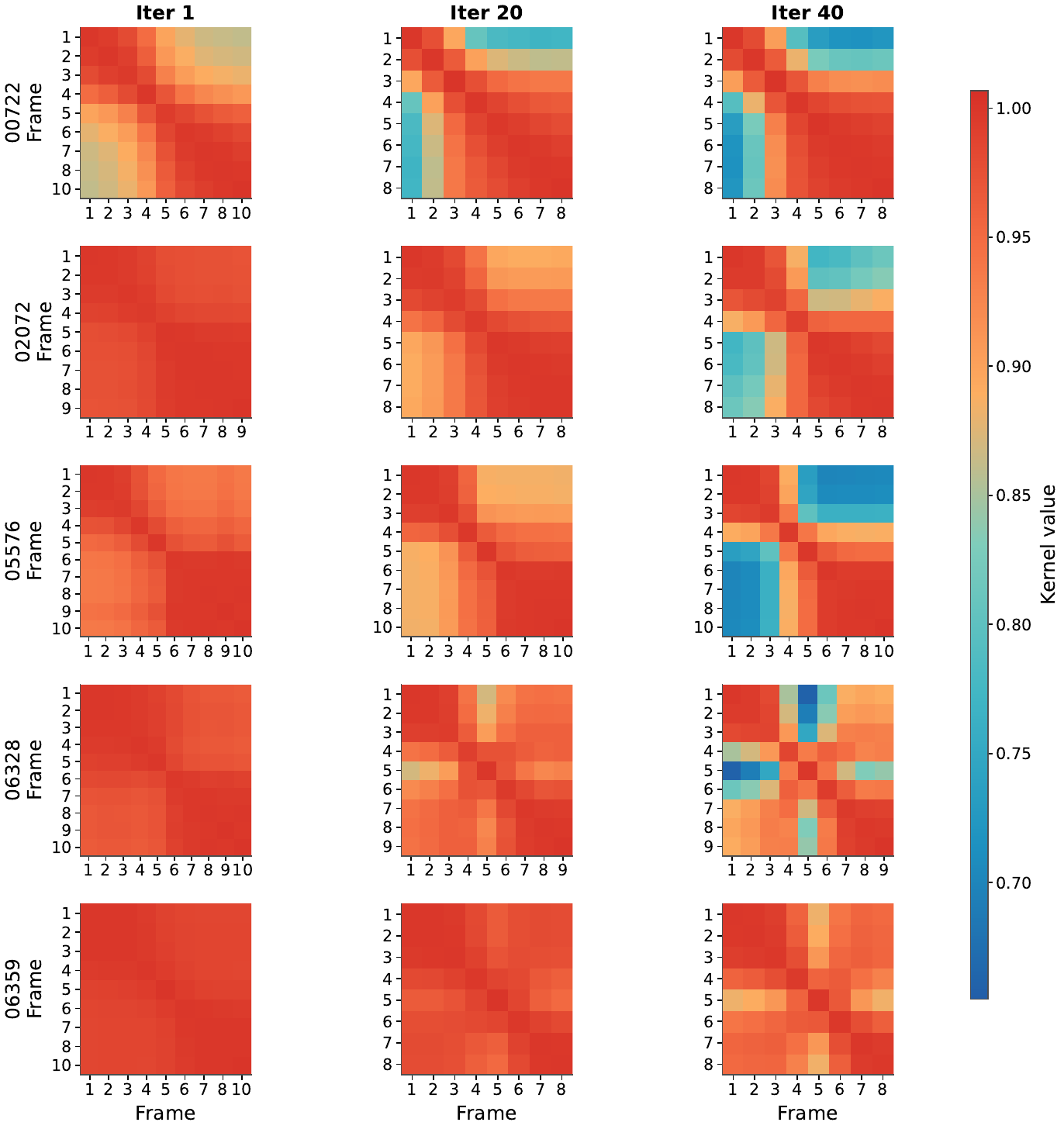}
  \captionof{figure}{Within-reaction scratch NTK frame-block kernels for the T1x subset after
  selected active-learning iterations. Each block orders structures by frame
  index along a reaction pathway.}
  \label{fig:scratchntk-frame-block-comparison}
\end{center}

\clearpage

\textbf{Pretrained Activation Kernels}

\begin{center}
  \centering
  \includegraphics[width=\linewidth]{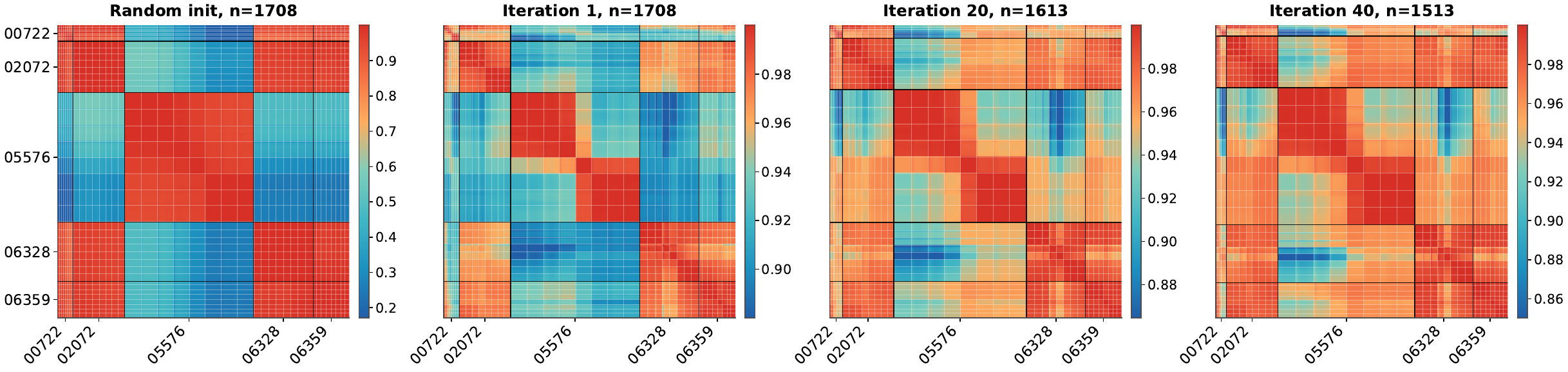}
  \captionof{figure}{Global activation-kernel matrices on the T1x subset, with structures
  ordered by reaction family. The panels compare random-initialized and
  active-learning iteration checkpoints using pooled scalar MACE activations.}
  \label{fig:activation-global-comparison}
\end{center}

\begin{center}
  \centering
  \includegraphics[width=\linewidth]{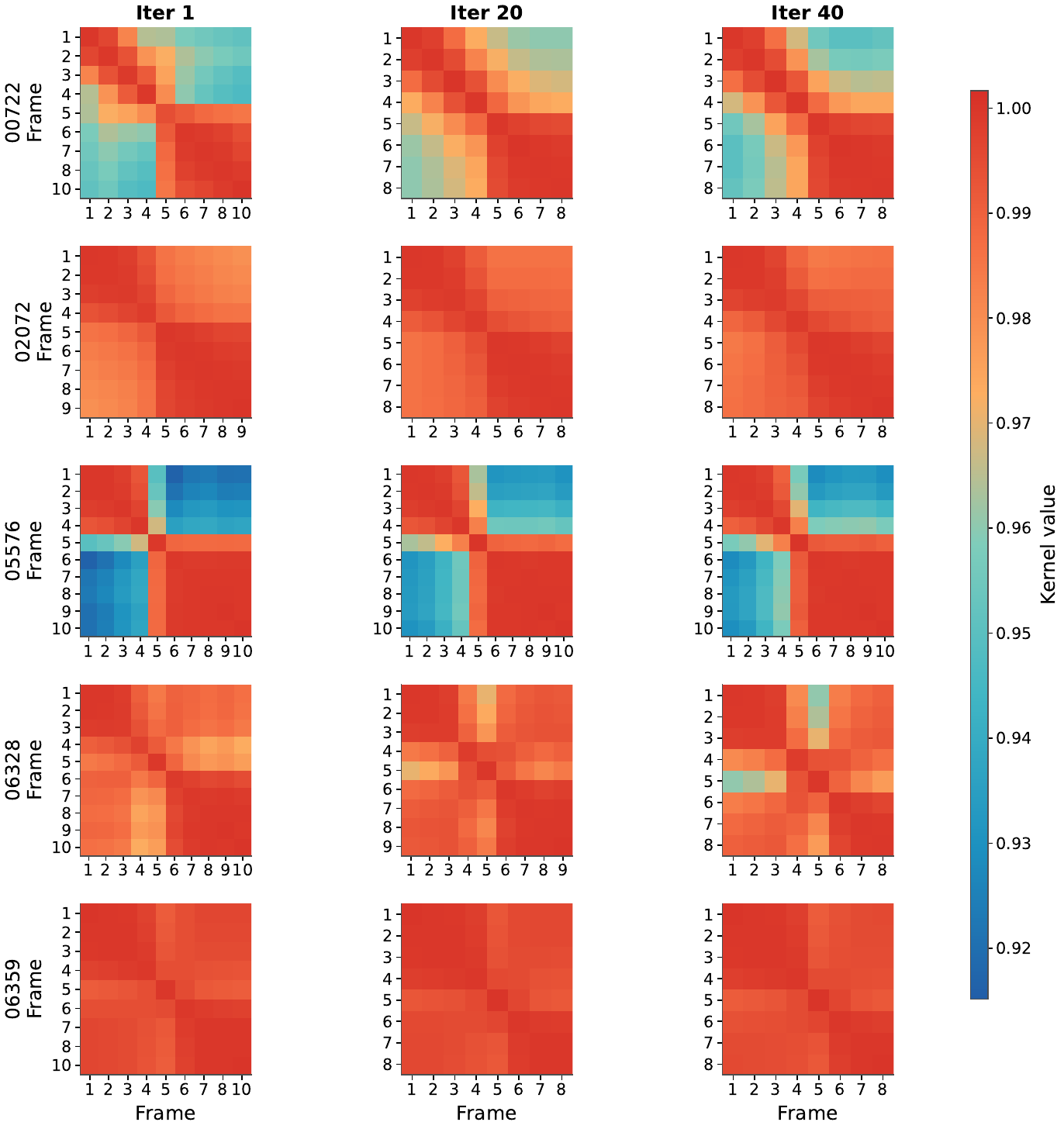}
  \captionof{figure}{Within-reaction activation frame-block kernels for the T1x subset
  after selected active-learning iterations.}
  \label{fig:activation-frame-block-comparison}
\end{center}

\clearpage
\textbf{Scratch Activation Kernels}

\begin{center}
  \centering
  \includegraphics[width=\linewidth]{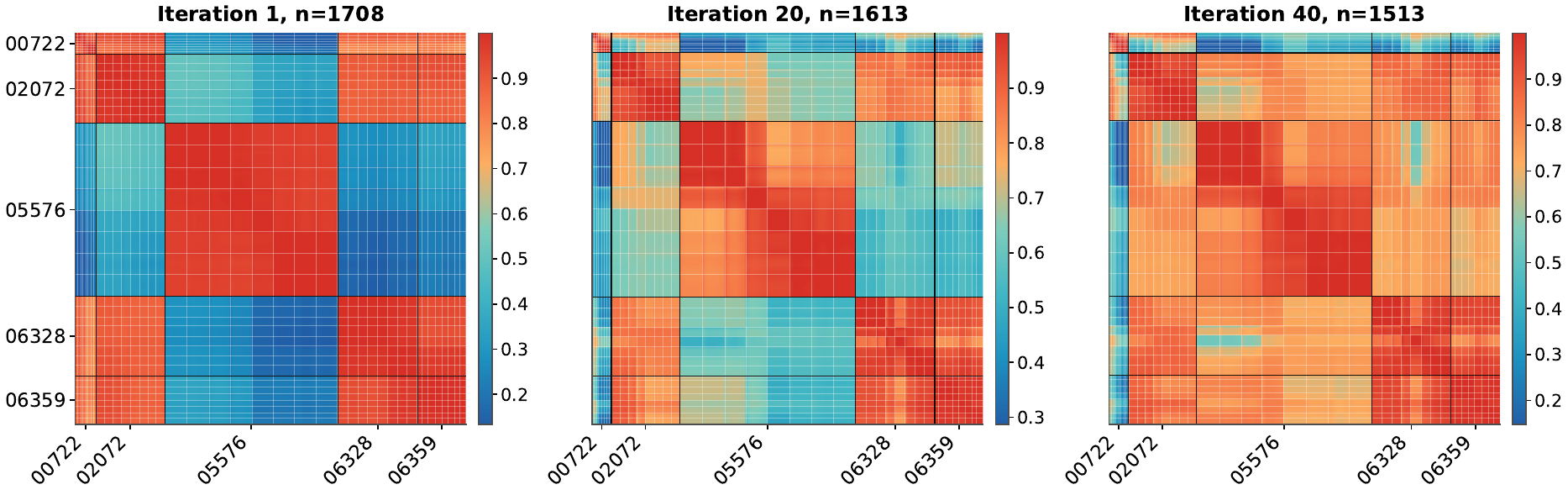}
  \captionof{figure}{Global srctach activation kernel matrices on the T1x subset, with structures
  ordered by reaction family.  The panels show how an untrained activation kernel evolves with AL iteration.}
  \label{fig:scratchactivation-global-comparison}
\end{center}

\begin{center}
  \centering
  \includegraphics[width=\linewidth]{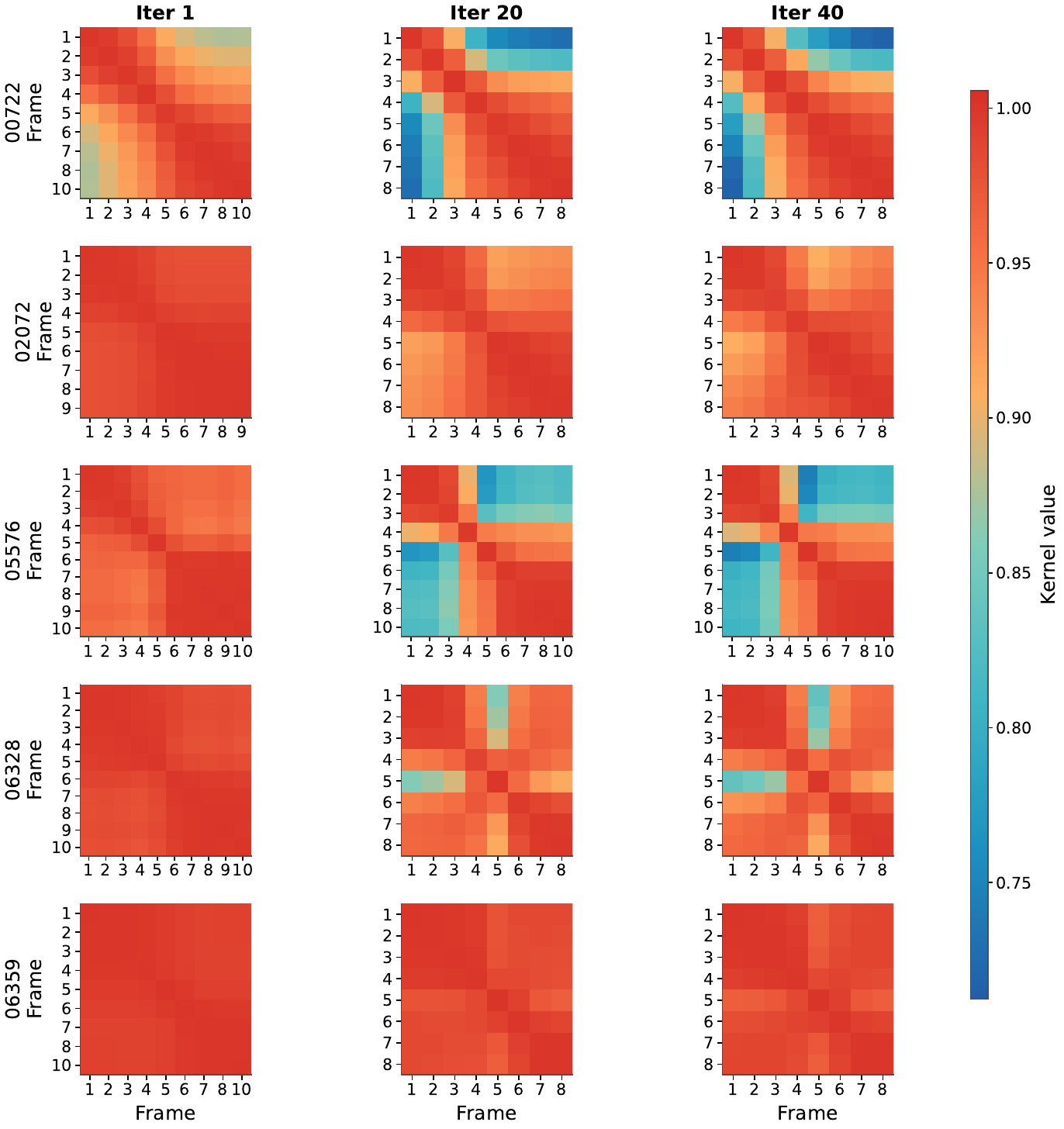}
  \captionof{figure}{Within-reaction scratch activation frame-block kernels for the T1x subset
  after selected active-learning iterations.}
  \label{fig:scratchactivation-frame-block-comparison}
\end{center}

\clearpage

\textbf{Descriptor Kernels}

\begin{center}
  \centering
  \begin{minipage}[t]{0.36\textwidth}
    \centering
    \includegraphics[width=\linewidth]{version1_figs/paper_plots/kernel_plots_t1x/soap_global_kernel.pdf}
    \centerline{(a) SOAP global kernel}
  \end{minipage}
  \hfill
  \begin{minipage}[t]{0.58\textwidth}
    \centering
    \includegraphics[width=\linewidth]{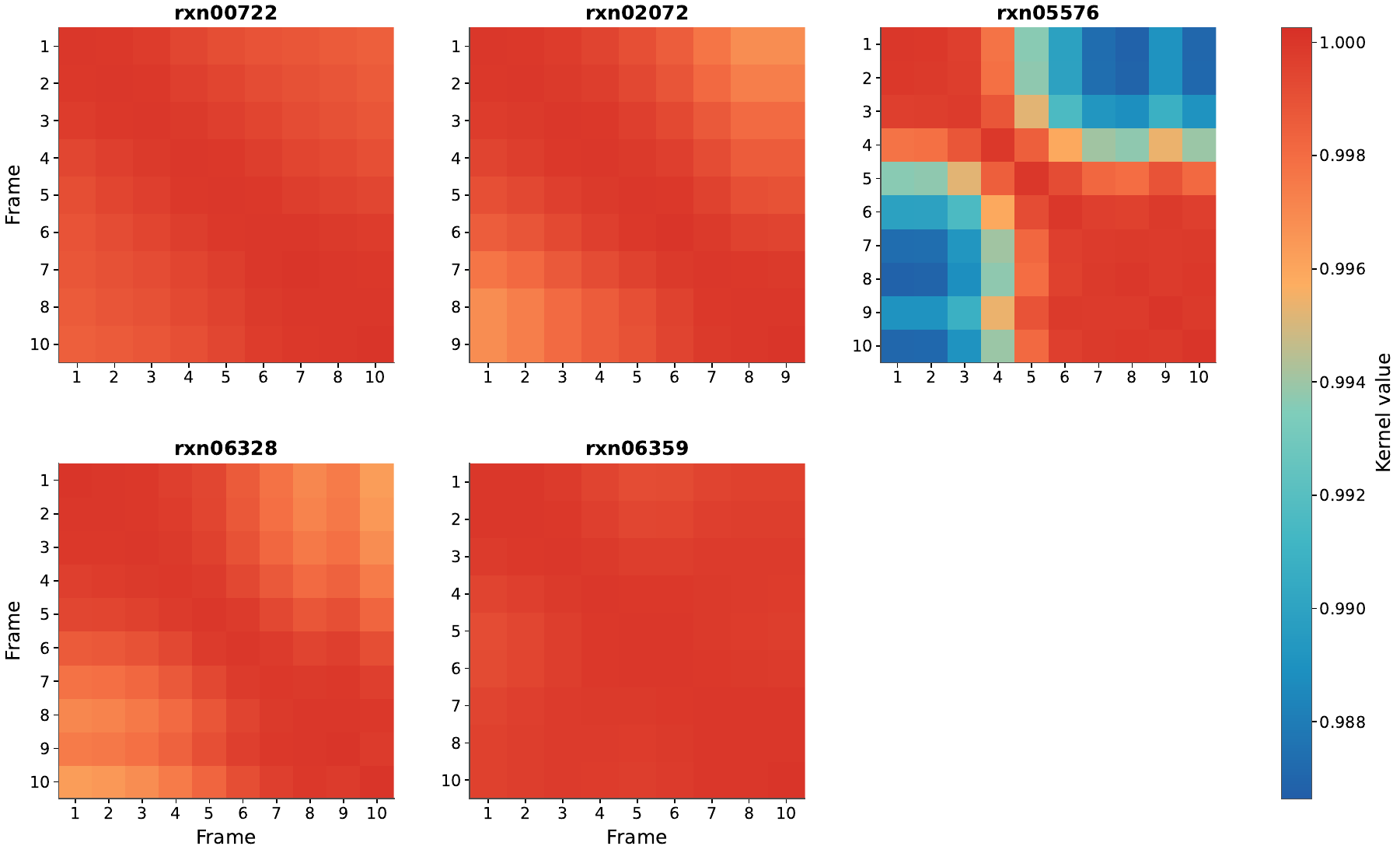}
    \centerline{(b) SOAP within-reaction blocks}
  \end{minipage}
  \captionof{figure}{SOAP kernel diagnostics on the T1x subset. SOAP captures coarse
  reaction-family structure using fixed local-geometry descriptors, but many
  within-reaction similarities remain high.}
  \label{fig:soap-detailed-kernels}
\end{center}

\begin{center}
  \centering
  \begin{minipage}[t]{0.36\textwidth}
    \centering
    \includegraphics[width=\linewidth]{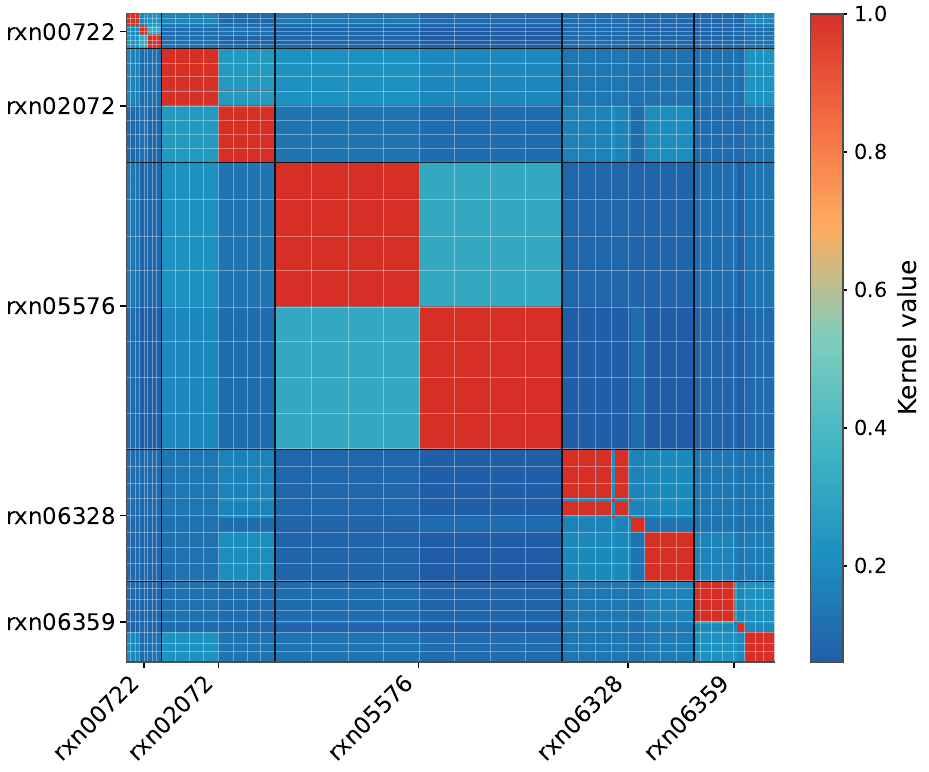}
    \centerline{(a) Tanimoto global kernel}
  \end{minipage}
  \hfill
  \begin{minipage}[t]{0.58\textwidth}
    \centering
    \includegraphics[width=\linewidth]{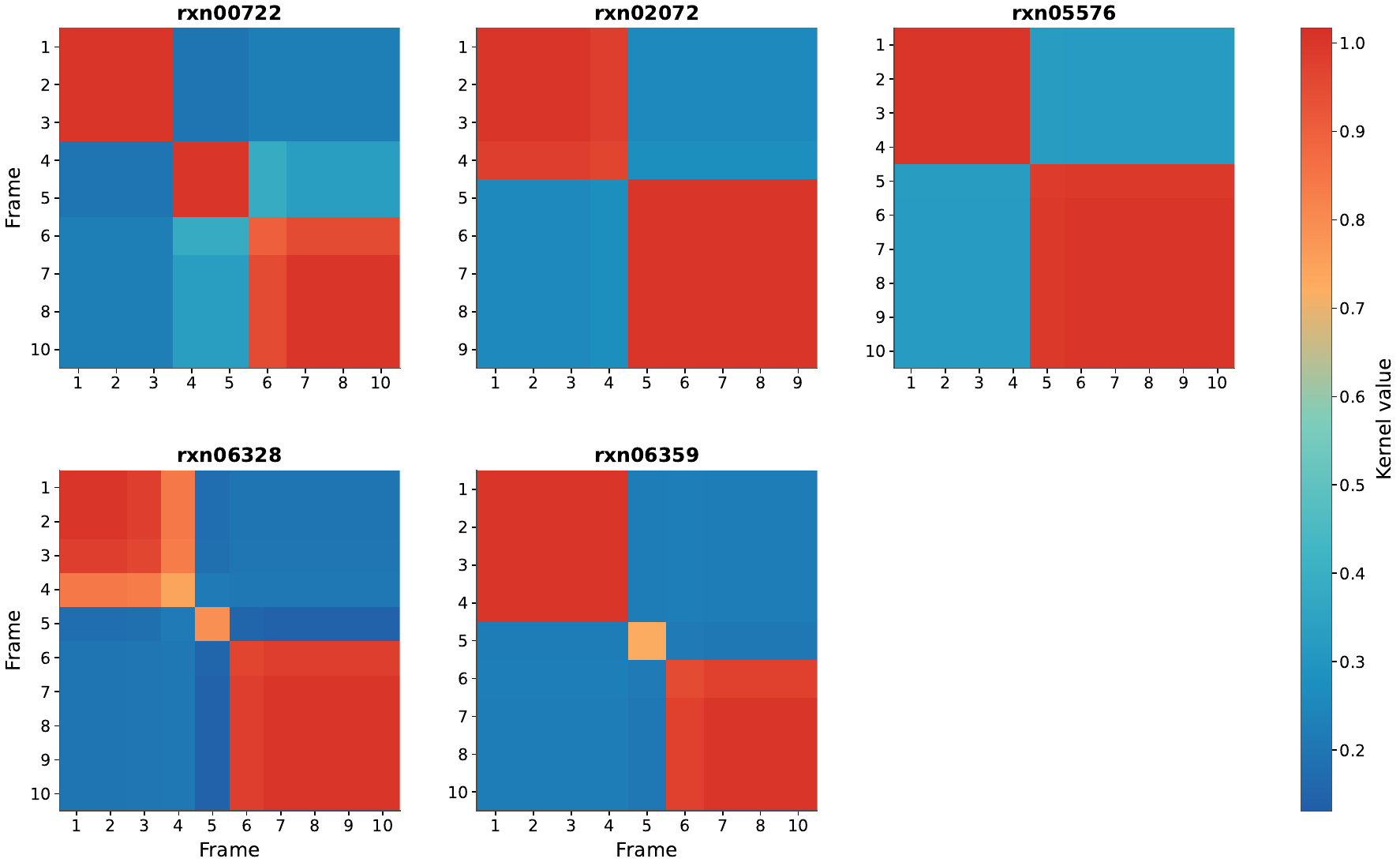}
    \centerline{(b) Tanimoto within-reaction blocks}
  \end{minipage}
  \captionof{figure}{Tanimoto kernel diagnostics on the T1x subset. Morgan fingerprints
  mainly reflect molecular graph identity and are less sensitive to continuous
  geometry changes along a fixed reaction path and are not able to capture inter reaction similarities.}
  \label{fig:tanimoto-detailed-kernels}
\end{center}

\clearpage
\subsection{Residual-GP Calibration}\label{app:residual_calibration}
For the residual GP experiment, for each kernel, we hold the pretrained MACE
  prediction $b(\mathbf{x})$ fixed and fit a Gaussian process to the residual
  target
  \[
    r(\mathbf{x}) = y - b(\mathbf{x}).
  \]
  Given a selected training set $\mathcal{T}$ and held-out structure
  $\mathbf{x}_i$, we center the training residuals by
  $\bar r_{\mathcal{T}} = |\mathcal{T}|^{-1}\sum_{j\in\mathcal{T}} r_j$ and use
  fixed kernel regularisation $\lambda=10^{-3}$. The residual-GP
  predictive correction and raw latent variance are
  \begin{align}
    \mu_i &=
    \bar r_{\mathcal{T}}
    +
    k_{\mathcal{T}}(\mathbf{x}_i)^\top
    (K_{\mathcal{T}\mathcal{T}}+\lambda I)^{-1}
    (\mathbf{r}_{\mathcal{T}}-\bar r_{\mathcal{T}}\mathbf{1}), \\
    s_i^2 &=
    k(\mathbf{x}_i,\mathbf{x}_i)
    -
    k_{\mathcal{T}}(\mathbf{x}_i)^\top
    (K_{\mathcal{T}\mathcal{T}}+\lambda I)^{-1}
    k_{\mathcal{T}}(\mathbf{x}_i).
  \end{align}
  The corrected prediction is
  \[
    \hat{y}_i = b(\mathbf{x}_i)+\mu_i.
  \]
  Since the embedding or similarity definition fixes the kernel geometry but not
  the residual signal scale, we estimate a per-prefix variance scale from the
  training residuals,
  \begin{equation}
    \hat{\gamma}
    =
    \frac{
    (\mathbf{r}_{\mathcal{T}}-\bar r_{\mathcal{T}}\mathbf{1})^\top
    (K_{\mathcal{T}\mathcal{T}}+\lambda I)^{-1}
    (\mathbf{r}_{\mathcal{T}}-\bar r_{\mathcal{T}}\mathbf{1})
    }{|\mathcal{T}|}.
  \end{equation}
  Gaussian negative log likelihood, predictive intervals, and calibration metrics
  are computed using predictive variance $\hat{\gamma}s_i^2$.

 We compare six residual kernels: pretrained activation features,
  randomly initialised activation features, pretrained NTK, randomly initialized
  NTK, SOAP, and Tanimoto. The randomly initialized neural kernels use the same
  MACE architecture before pretraining, separating architectural inductive bias
  from representation structure learned during pretraining.

For each kernel, we replay posterior-variance acquisition from the same initial seed to produce a nested sequence of training prefixes
$\mathcal{T}_1 \subset \mathcal{T}_2 \subset \cdots$. Each prefix is then used to fit a residual Gaussian process (GP). Because calibration can change as additional data is acquired, we select the calibrated GP on the validation split using the Gaussian negative log-likelihood (NLL).

\clearpage
\section{Additional Datasets Plots}
\label{sec:appendix-transferability}

\begin{figure}[H]
    \centering
    \includegraphics[width=\linewidth]{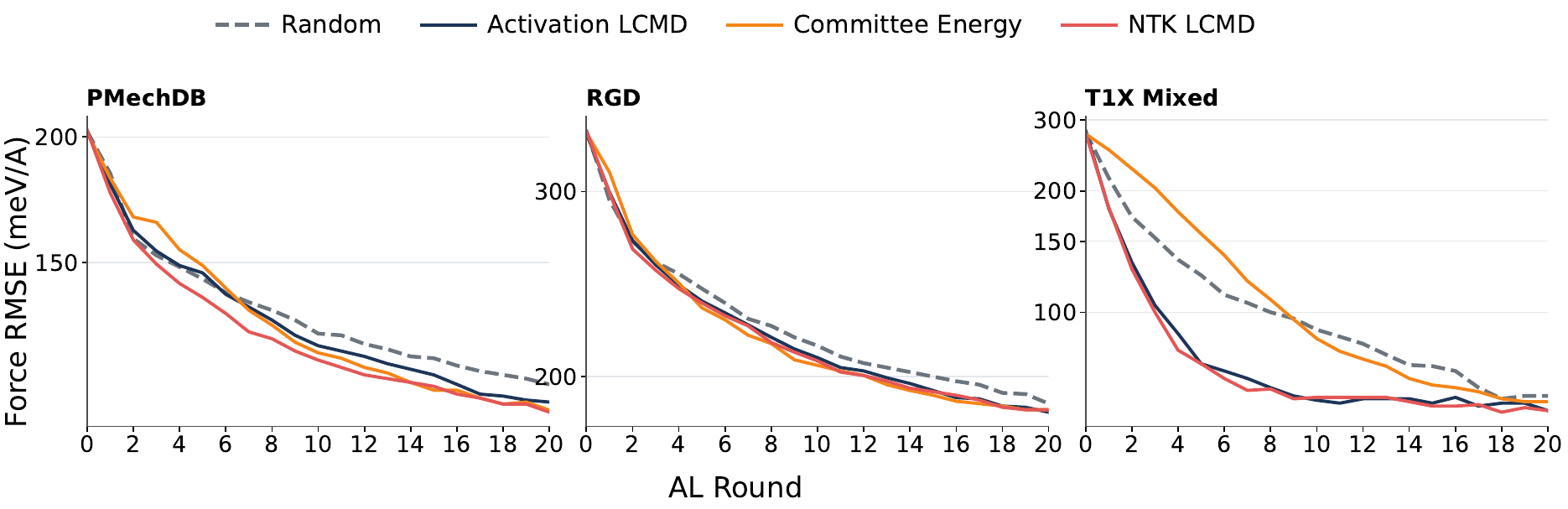}
    \caption{Additional transferability results on PMechDB, RGD, and T1x Mixed. The curves show force RMSE across active-learning rounds.}
    \label{fig:transfer_force_rmse}
\end{figure}

\begin{figure}[H]
    \centering
    \includegraphics[width=\linewidth]{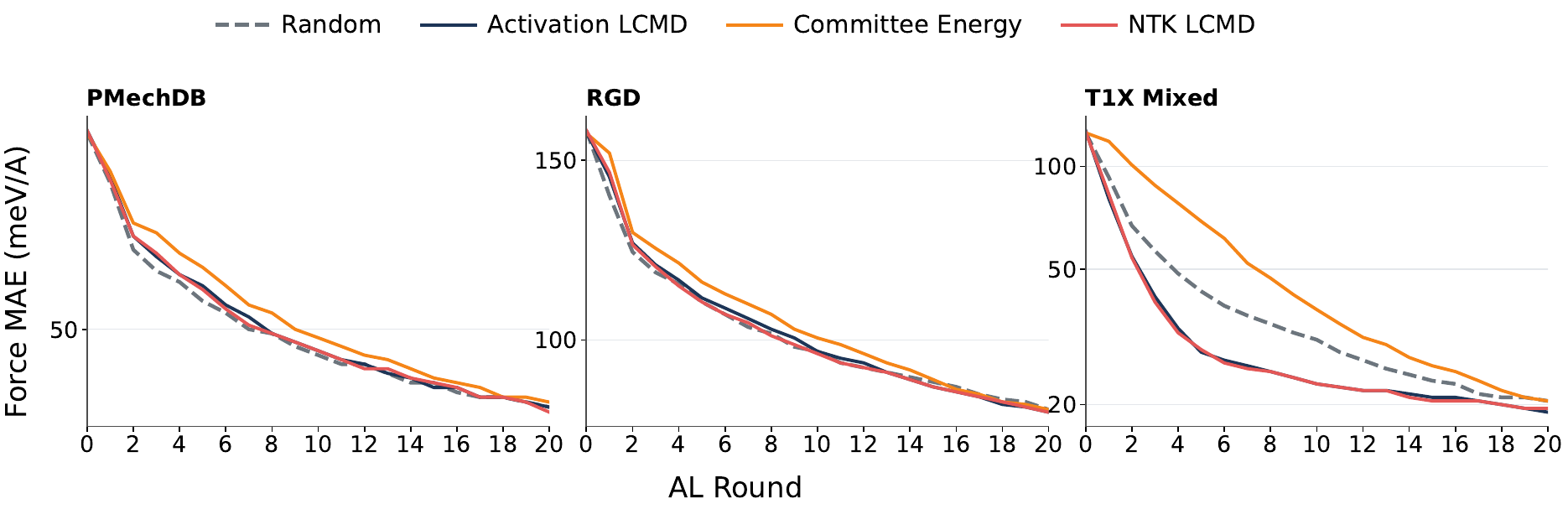}
    \caption{Additional transferability results on PMechDB, RGD, and T1x Mixed. The curves show force MAE across active-learning rounds.}
    \label{fig:transfer_force_mae}
\end{figure}

\begin{figure}[H]
    \centering
    \includegraphics[width=\linewidth]{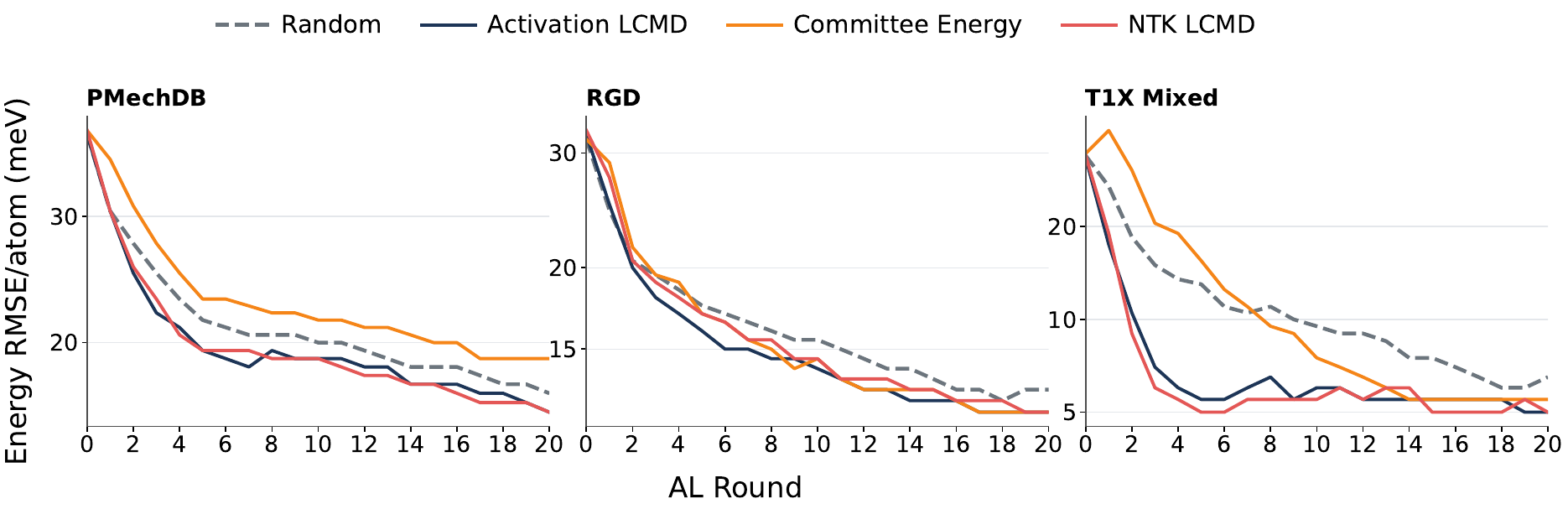}
    \caption{Additional transferability results on PMechDB, RGD, and T1x Mixed. The curves show energy RMSE across active-learning rounds.}
    \label{fig:transfer_energy_rmse}
\end{figure}

\begin{figure}[H]
    \centering
    \includegraphics[width=\linewidth]{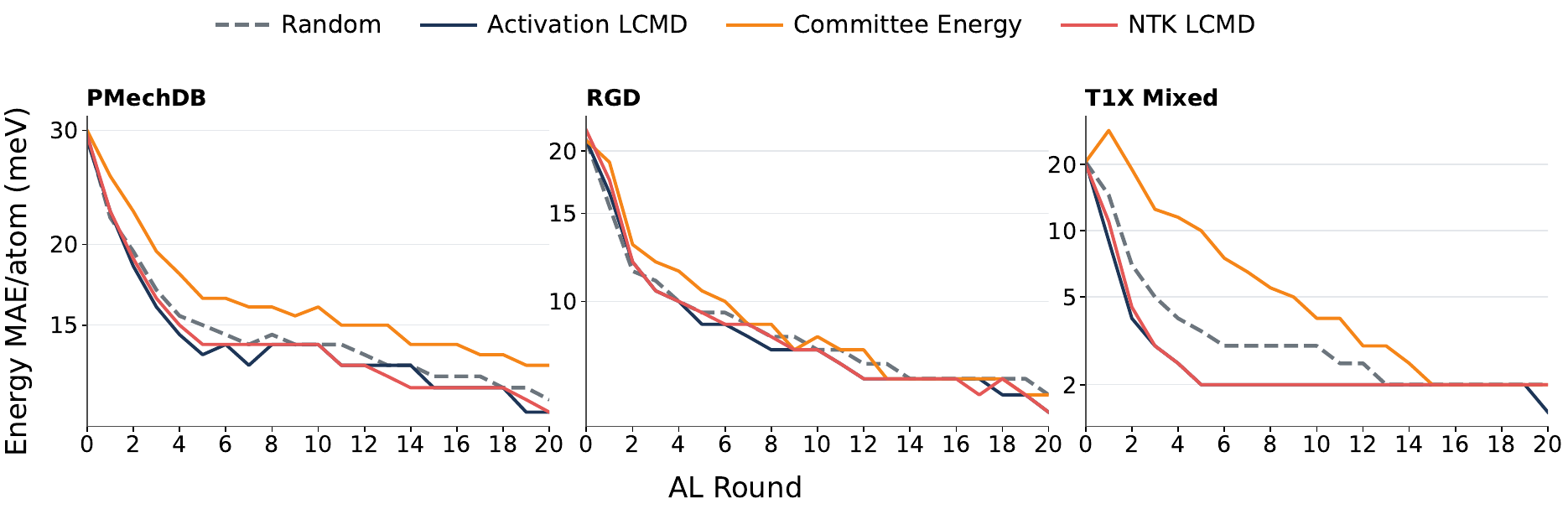}
    \caption{Additional transferability results on PMechDB, RGD, and T1x Mixed. The curves show energy MAE across active-learning rounds.}
    \label{fig:transfer_energy_mae}
\end{figure}

This appendix provides the detailed learning curves underlying the round-gain summary in Table~\ref{tab:transfer_benchmarks}. For each of the three datasets, PMechDB, RGD, and T1x Mixed, we report energy and force errors in both RMSE and MAE form across acquisition rounds, comparing model-based kernels (Activation-LCMD and NTK-LCMD), committee energy disagreement, and random acquisition. Figure~\ref{fig:transfer_force_rmse} shows force RMSE, Figure~\ref{fig:transfer_force_mae} shows force MAE, Figure~\ref{fig:transfer_energy_rmse} shows energy RMSE, and Figure~\ref{fig:transfer_energy_mae} shows energy MAE.

\end{document}